\documentclass[usenames,dvipsnames]{article} % For LaTeX2
\usepackage{iclr2019_conference,times}

\usepackage{amsmath}
\usepackage{amsthm}
\usepackage{amsfonts}
\newcommand{\norm}[1]{\left\lVert#1\right\rVert}

\newtheorem{prop}{Proposition}

\usepackage{bm}
\usepackage{xcolor}

\usepackage{url}
\usepackage{algorithm}
\usepackage[noend]{algorithmic}
\usepackage{graphicx}
\renewcommand{\algorithmiccomment}[1]{\bgroup\hfill $\triangleright$ ~#1\egroup}
\usepackage{hyperref}

\title{Decoupled Weight Decay Regularization}

\newcommand{\franknips}[1]{\textcolor{black}{#1}}

\def\R{{\rm I\hspace{-0.50ex}R}}
\def\E{\mathds{E}}
\def\D{{\mathcal{D}}}
\def\H{\textbf{H}}
\def\A{\cal{A}}
\newcommand{\vc}[1]{\textit{\textbf{#1}}}

\iclrfinalcopy

\author{Ilya Loshchilov \& Frank Hutter \\
University of Freiburg\\
Freiburg, Germany, \\
\texttt{\{ilya,fh\}@cs.uni-freiburg.de} 
}

\begin{document}

\maketitle

\begin{abstract} 
L$_2$ regularization and weight decay regularization are equivalent for standard stochastic gradient descent (when rescaled by the learning rate), but as we demonstrate this is \emph{not} the case for adaptive gradient algorithms, such as Adam.
While common implementations of these algorithms employ L$_2$ regularization (often calling it ``weight decay'' in what may be misleading due to the inequivalence we expose), we propose a simple modification to recover the original formulation of weight decay regularization by \emph{decoupling} the weight decay from the optimization steps taken w.r.t. the loss function.
We provide empirical evidence that our proposed modification (i) decouples the optimal choice of weight decay factor from the setting of the learning rate for both standard SGD and Adam and (ii) substantially improves Adam's generalization performance, allowing it to compete with SGD with momentum on image classification datasets (on which it was previously typically outperformed by the latter).
Our proposed decoupled weight decay has already been adopted by many researchers, and the community has implemented it in TensorFlow and PyTorch; the complete source code for our experiments is available at \url{https://github.com/loshchil/AdamW-and-SGDW}

%We note that common implementations of adaptive gradient algorithms, such as Adam, limit the potential benefit of weight decay regularization, because the weights do not decay multiplicatively (as would be expected for standard weight decay) but by an additive constant factor. 
%We propose a simple way to resolve this issue by decoupling weight decay and the optimization steps taken w.r.t. the loss function. We provide empirical evidence that our proposed modification (i) decouples the optimal choice of weight decay factor from the setting of the learning rate for both standard SGD and Adam, and (ii) substantially improves Adam's generalization performance, allowing it to compete with SGD with momentum on image classification datasets (on which it was previously typically outperformed by the latter).
%We also demonstrate that longer optimization runs require smaller weight decay values for optimal results and introduce a normalized variant of weight decay to reduce this dependence. Finally, we propose a version of Adam with warm restarts (AdamWR) that has strong anytime performance while achieving state-of-the-art results on CIFAR-10 and ImageNet32x32. 
%Our source code will become available after the review process.
%Our source code is available at \url{https://github.com/loshchil/AdamW-and-SGDW}
\end{abstract}

%%%%%%%%%%%%%%%%%%%%%%%%%%%%%%%%%%%%%%%%%%%%%%%%%%%%%%%%%%%%%%%%%%%%%%%%%%%%%%%%%%%%%%%
\section{Introduction}
%%%%%%%%%%%%%%%%%%%%%%%%%%%%%%%%%%%%%%%%%%%%%%%%%%%%%%%%%%%%%%%%%%%%%%%%%%%%%%%%%%%%%%%

\def\R{{\rm I\hspace{-0.50ex}R}}
\def\E{\mathds{E}}
\def\D{{\mathcal{D}}}
\def\H{\textbf{H}}
\def\A{\cal{A}}

\newcommand{\ma}[1]{\mathchoice{\mbox{\boldmath$\displaystyle#1$}}
  {\mbox{\boldmath$\textstyle#1$}} {\mbox{\boldmath$\scriptstyle#1$}}
  {\mbox{\boldmath$\scriptscriptstyle#1$}}}
\renewcommand{\ma}[1]{\mathnormal{\mathbf{#1}}}
\newcommand{\mstr}[1]{\mathrm{#1}}
\newcommand{\C}{ \ensuremath{\ma{C}} }
\newcommand{\I}{ \ensuremath{\ma{I}} }
\newcommand{\M}{ \ensuremath{\ma{M}} }
\newcommand{\NormalNullC}{{\mathcal N}  \hspace{-0.13em}\left({\ma{0},\C\,}\right)}
\newcommand{\dd}{n}

\def\UU{{\rm I\hspace{-0.50ex}U}}

\def\RR{{\rm I\hspace{-0.50ex}R}}

\def\NormOI{{\mathcal N}  \hspace{-0.13em}\left({\ma{0}, \ensuremath{\ma{I}}\,}\right)}
\def\ONE{{\rm 1\hspace{-0.80ex}1}}
\def\Id{\ensuremath{\ma{I}}}
\def\MYUNDERLINE{ $\noindent\underline{\makebox[0.06in][l]{}}$ }
\def\x{\bm{\theta}}
\def\y{\vc{y}}
\def\m{\vc{m}}
\def\vy{\vc{y}}
\newcommand{\HYP}{H}
\def\UU{{\rm I\hspace{-0.60ex}U}}

% FH old: I radically cut text to get to the point right away. I've left the original source commented below.
%
%The training of a Deep Neural Network (DNN) with $n$ free parameters can be formulated as the problem of minimizing a function $f: \R^n \rightarrow \R$. The commonly used procedure to optimize $f$ is to iteratively adjust $\bm{\theta}_t \in \R^n$ (the parameter vector at time step $t$) using gradient information $\nabla f_t(\bm{\theta}_t)$ obtained on a small $t$-th batch of $b$ out of $B$ datapoints. The Stochastic Gradient Descent (SGD) procedure then becomes an extension of the Gradient Descent to stochastic optimization of $f$:
%\begin{eqnarray}
%	\label{eq:sgd}
%	\bm{\theta}_{t+1} = \bm{\theta}_t - \alpha \nabla f_t(\bm{\theta}_t),
%\end{eqnarray}
%
%where $\alpha$ is a learning rate. Basic extensions of SGD such as SGD with momentum defined as
%
%\begin{eqnarray}
%	\vc{m}_{t+1} = \mu_t \vc{m}_t  - \alpha \nabla f_t(\bm{\theta}_t), \label{eq:moms1} \\
%	\bm{\theta}_{t+1} = \bm{\theta}_t + \vc{m}_{t+1}, \label{eq:moms2}
%\end{eqnarray}
%
%represent a common baseline method for training DNNs. 

Adaptive gradient methods, such as AdaGrad~\citep{duchi2011adaptive}, RMSProp~\citep{tieleman2012lecture}, Adam~\citep{kingma2014adam} and most recently AMSGrad ~\citep{reddi2018iclr} have become a default method of choice for training feed-forward and recurrent neural networks \citep{xu2015show,  radford2015unsupervised}. Nevertheless, state-of-the-art results for popular image classification datasets, such as CIFAR-10 and CIFAR-100~\cite{krizhevsky2009learning}, are still obtained by applying SGD with momentum \citep{gastaldi2017shake, cubuk2018autoaugment}. Furthermore, \cite{wilson2017marginal} suggested that adaptive gradient methods do not generalize as well as SGD with momentum when tested on a diverse set of deep learning tasks, such as image classification, character-level language modeling and constituency parsing.
Different hypotheses about the origins of this worse generalization have been investigated, such as the presence of sharp local minima~\citep{keskar2016large,dinh2017sharp} and inherent problems of adaptive gradient methods \citep{wilson2017marginal}. In this paper, we investigate whether it is better to use L$_2$ regularization or weight decay regularization to train deep neural networks with SGD and Adam. We show that a major factor of the poor generalization of the most popular adaptive gradient method, Adam, is due to the fact that L$_2$ regularization is not nearly as effective for it as for SGD. %We also demonstrate how to resolve this problem. %that this problem can be resolved by using the original formulation of weight decay.
Specifically, our analysis of Adam %given in this paper 
leads to the following observations: 
% This issue also pertains to other adaptive gradient methods.

%Interestingly, many state-of-the-art results for popular image classification datasets such as CIFAR-10 and CIFAR-100 are obtained by applying SGD with momentum \citep{SnapshotICLR2017, huang2016densely, loshchilov2016sgdr,gastaldi2017shake}. 
%This observation is in contrast with the growing popularity of the Adam algorithm \citep{kingma2014adam} as the method of choice for training feed-forward and recurrent DNNs \citep{xu2015show, gregor2015draw, kumar2015ask, radford2015unsupervised}. \cite{wilson2017marginal} suggested that adaptive gradient methods such as Adam, AdaGrad, RMSProp do not generalize as well as SGD with momentum when tested on a diverse set of deep learning tasks such as image classification, character-level language modeling and constituency parsing. 

\begin{description}
	\item \textbf{L$_2$ regularization and weight decay are not identical.} The two techniques can be made equivalent for SGD by a reparameterization of the weight decay factor based on the learning rate; however, as is often overlooked, this is not the case for Adam. 
In particular, when combined with adaptive gradients, L$_2$ regularization leads to weights with large historic parameter and/or gradient amplitudes being regularized less than they would be when using weight decay.

	\item \textbf{L$_2$ regularization is not effective in Adam.} One possible explanation why Adam and other adaptive gradient methods might be outperformed by SGD with momentum is that common deep learning libraries only implement L$_2$ regularization, not the original weight decay. Therefore, on tasks/datasets where the use of L$_2$ regularization is beneficial for SGD (e.g., on many popular image classification datasets), Adam leads to worse results than SGD with momentum (for which L$_2$ regularization behaves as expected). 
    
	\item \textbf{Weight decay
    %The original formulation of  
    %works as expected and 
    is equally effective in both SGD and Adam.} For SGD, it is equivalent to L$_2$ regularization, while for Adam it is not.
    
	\item \textbf{Optimal weight decay depends on the total number of batch passes/weight updates.} Our empirical analysis of SGD and Adam suggests that the larger the runtime/number of batch passes to be performed, the smaller the optimal weight decay. 
%This effect tends to be neglected because hyperparameters are often tuned for a fixed 
%%or a comparable 
%number of training epochs. As a result, the values of the weight decay found to perform best for short runs do not generalize to much longer runs. 

	\item \textbf{Adam can substantially benefit from a scheduled learning rate multiplier.}
    The fact that Adam is an adaptive gradient algorithm and as such adapts the learning rate for each parameter does \emph{not} rule out the possibility to substantially improve its performance by using a global learning rate multiplier, scheduled, e.g., by cosine annealing.
\end{description}
 
The main contribution of this paper is to \emph{improve regularization in Adam by decoupling the weight decay from the gradient-based update}. In a comprehensive analysis, we show that Adam generalizes substantially better with decoupled weight decay than with L$_2$ regularization, achieving 15\% relative improvement in test error (see Figures \ref{fig2_hyp100epochs} and \ref{fig1800}); this holds true for various image recognition datasets (CIFAR-10 and ImageNet32x32), training budgets (ranging from 100 to 1800 epochs), and learning rate schedules (fixed, drop-step, and cosine annealing; see Figure \ref{fig:adam_with_without_cosine_stepdrop}).
We also demonstrate that our decoupled weight decay renders the optimal settings of the learning rate and the weight decay factor much more independent, thereby easing hyperparameter optimization (see Figure \ref{fig2_hyp100epochs}).

The main motivation of this paper is to improve Adam 
%by means of the original formulation weight decay in Adam 
to make it competitive w.r.t.\ SGD with momentum even for those problems where it did not use to be competitive. 
%We also improve Adam's practical performance based on cosine annealing and warm restarts, which had previously only been demonstrated to be efficient for SGD. % with momentum.
We hope that as a result, practitioners do not need to switch between Adam and SGD anymore, which in turn should reduce the common issue of selecting dataset/task-specific training algorithms and their hyperparameters.

%In this paper, we argue that the currently common way of implementing the weight decay regularization in deep learning might limit potential benefits that the weight decay can deliver. More specifically, we suggest that instead of employing Eq. (\ref{eq:freq}) or Eq. (\ref{eq:freqgrad}), one should decay the weights separately from the update based on $\nabla f(\bm{\theta})$. 

\definecolor{newcolor}{rgb}{0.8,1,1}

\newcommand{\adamcolor}{Thistle}
\newcommand{\ouradamcolor}{SpringGreen}
\newcommand{\adam}[1]{\colorbox{\adamcolor}{$\displaystyle #1$}}
\newcommand{\adamtext}[1]{\colorbox{\adamcolor}{#1}}
\newcommand{\ouradam}[1]{\colorbox{\ouradamcolor}{$\displaystyle #1$}}
\newcommand{\ouradamtext}[1]{\colorbox{\ouradamcolor}{#1}}

\section{Decoupling the Weight Decay from the Gradient-based Update}
\label{sec:decoupling}

In the weight decay described by \citet{hanson1988comparing}, the weights $\bm{\theta}$  decay exponentially as 
\begin{eqnarray}
	\bm{\theta}_{t+1} = (1 - \lambda) \bm{\theta}_t - \alpha \nabla f_t(\bm{\theta}_t), \label{eq:wdecay}
\end{eqnarray}
where $\lambda$ defines the rate of the weight decay per step and $\nabla f_t(\bm{\theta}_t)$ is the $t$-th batch gradient to be multiplied by a learning rate $\alpha$.
For standard SGD, %this mechanism 
it is equivalent to standard L$_2$ regularization: 
\begin{prop}[Weight decay = L$_2$ reg for standard SGD]
Standard SGD with base learning rate $\alpha$ executes the same steps on batch loss functions $f_t(\bm{\theta})$ with weight decay $\lambda$ (defined in Equation \ref{eq:wdecay}) as it executes without weight decay on $f_{t}^{\text{reg}}(\bm{\theta}) = f_t(\bm{\theta}) + \frac{\lambda'}{2} \norm{\bm{\theta}}_2^2$, with $\lambda' = \frac{\lambda}{\alpha}$.
\end{prop}
The proofs of this well-known fact, as well as our other propositions, are given in Appendix \ref{sec:decay_vs_L_2}.

Due to this equivalence, L$_2$ regularization is very frequently referred to as weight decay, including in popular deep learning libraries. However, as we will demonstrate later in this section, this equivalence does \emph{not} hold for adaptive gradient methods. 
One fact that is often overlooked already for the simple case of SGD is that in order for the equivalence to hold, the L$_2$ regularizer $\lambda'$ has to be set to $\frac{\lambda}{\alpha}$, i.e., if there is an overall best weight decay value $\lambda$, the best value of $\lambda'$ is tightly coupled with the learning rate $\alpha$. 
In order to decouple the effects of these two hyperparameters, we advocate to decouple the weight decay step as proposed by \citet{hanson1988comparing} (Equation \ref{eq:wdecay}).

%Following \citet{hanson1988comparing}, 
%one can also modify the original batch loss $f_t(\bm{\theta}_t)$ and consider a bias term (also referred to as the regularization term) accounting for ``costs'' on weights which are, e.g., quadratic in the weight values as for L$_2$  regularization: 
%\begin{eqnarray}
%	f_{t}^{\text{reg}}(\bm{\theta}_t) = f_t(\bm{\theta}_t) + \frac{\lambda}{2}  \left\| \bm{\theta}_t \right\|^2_2, \label{eq:freq}
%\end{eqnarray}
%
%where $\lambda$ defines the impact of the L$_2$  regularization. In order to consider the weight decay regularization, one can reformulate the objective function as in Eq. (\ref{eq:freq}) or directly adjust $\nabla f_t(\bm{\theta}_t)$ as
%
%\begin{eqnarray}
%	\nabla f_{t}^{\text{reg}}(\bm{\theta}_t) = \nabla f_t(\bm{\theta}_t) + \lambda\bm{\theta}_t.  \label{eq:freqgrad}
%\end{eqnarray}

\begin{algorithm}[tb!]
\caption{\adamtext{SGD with L$_2$ regularization} and \ouradamtext{SGD with \franknips{decoupled} weight decay (SGDW)}, both with momentum}
\footnotesize
\label{algo_sgd}
\begin{algorithmic}[1]
\STATE{\textbf{given} initial learning rate $\alpha \in \R$, momentum factor $\beta_1 \in \R$, weight decay/L$_2$ regularization factor $\lambda \in \R$}\label{adam-Given}
\STATE{\textbf{initialize} time step $t \leftarrow 0$, parameter vector $\bm{\theta}_{t=0} \in \R^n$,  first moment vector $\vc{m}_{t=0} \leftarrow \vc{0}$, schedule multiplier $\eta_{t=0} \in \R$}
\REPEAT
	\STATE{$t \leftarrow t + 1$}
	\STATE{$\nabla f_t(\bm{\theta}_{t-1}) \leftarrow  \text{SelectBatch}(\bm{\theta}_{t-1})$}  \COMMENT{select batch and return the corresponding gradient}
	\STATE{$\vc{g}_t \leftarrow \nabla f_t(\bm{\theta}_{t-1})$  \adam{+ \lambda\bm{\theta}_{t-1}}} \label{sgd-computegrad}
	\STATE{$\eta_t \leftarrow \text{SetScheduleMultiplier}(t)$}	\COMMENT{can be fixed, decay, be used for warm restarts}
	\STATE{$\vc{m}_t \leftarrow \beta_1 \vc{m}_{t-1} + \eta_t \alpha \vc{g}_t $} \label{sgd-mom1} 
	\STATE{$\bm{\theta}_t \leftarrow \bm{\theta}_{t-1} - \vc{m}_t$ \ouradam{- \eta_t \lambda\bm{\theta}_{t-1}}}  \label{sgd-xupdate} 
\UNTIL{ \textit{stopping criterion is met} }
\RETURN{optimized parameters $\bm{\theta}_t$}
\end{algorithmic}
\end{algorithm}

\begin{algorithm}[tb!]
\caption{\adamtext{Adam with L$_2$ regularization} and \ouradamtext{Adam with \franknips{decoupled} weight decay (AdamW)}}
\footnotesize
\label{algo_adam}
\begin{algorithmic}[1]
\STATE{\textbf{given} $\alpha = 0.001, \beta_1 = 0.9, \beta_2 =0.999, \epsilon = 10^{-8}, \lambda\in \R$} \label{adam-Given}
\STATE{\textbf{initialize} time step $t \leftarrow 0$, parameter vector $\bm{\theta}_{t=0} \in \R^n$,  first moment vector $\vc{m}_{t=0} \leftarrow \vc{0}$, second moment vector  $\vc{v}_{t=0} \leftarrow \vc{0}$, schedule multiplier $\eta_{t=0} \in \R$}
\REPEAT
	\STATE{$t \leftarrow t + 1$}
	\STATE{$\nabla f_t(\bm{\theta}_{t-1}) \leftarrow  \text{SelectBatch}(\bm{\theta}_{t-1})$}  \COMMENT{select batch and return the corresponding gradient}
	\STATE{$\vc{g}_t \leftarrow \nabla f_t(\bm{\theta}_{t-1})$  \adam{+ \lambda\bm{\theta}_{t-1}}}
	\STATE{$\vc{m}_t \leftarrow \beta_1 \vc{m}_{t-1} + (1 - \beta_1) \vc{g}_t $} \label{adam-mom1} \COMMENT{here and below all operations are element-wise}
	\STATE{$\vc{v}_t \leftarrow \beta_2 \vc{v}_{t-1} + (1 - \beta_2) \vc{g}^2_t $} \label{adam-mom2}
	\STATE{$\hat{\vc{m}}_t \leftarrow \vc{m}_t/(1 - \beta_1^t) $} \COMMENT{$\beta_1$ is taken to the power of $t$} \label{adam-corr1}
	\STATE{$\hat{\vc{{v}}}_t \leftarrow \vc{v}_t/(1 - \beta_2^t) $} \COMMENT{$\beta_2$ is taken to the power of $t$} \label{adam-corr2}
	\STATE{$\eta_t \leftarrow \text{SetScheduleMultiplier}(t)$}	\COMMENT{can be fixed, decay, or also be used for warm restarts}
	\STATE{$\bm{\theta}_t \leftarrow \bm{\theta}_{t-1} - \eta_t \left( \alpha  \hat{\vc{m}}_t / (\sqrt{\hat{\vc{v}}_t} + \epsilon) \ouradam{+ \lambda\bm{\theta}_{t-1}} \right)$} \label{adam-xupdate}
\UNTIL{ \textit{stopping criterion is met} }
\RETURN{optimized parameters $\bm{\theta}_t$}
\end{algorithmic}
\end{algorithm}

%Historically, stochastic gradient descent methods inherited this way of implementing the weight decay regularization. 
%
%The currently most common way (e.g., in popular libraries such as TensorFlow, Keras, PyTorch, Torch, and Lasagne) 
%to introduce weight decay regularization is to use the L$_2$ regularization term as in Eq. (\ref{eq:freq}) or, often equivalently, to directly modify the gradient as in Eq. (\ref{eq:freqgrad}). 

Looking first at the case of SGD, we propose to decay the weights simultaneously with the update of $\bm{\theta}_t$ based on gradient information in Line 9 of Algorithm 1. This yields our proposed variant of SGD with momentum using decoupled weight decay (\textbf{SGDW}). 
This simple modification explicitly decouples $\lambda$ and $\alpha$ (although some problem-dependent implicit coupling may of course remain as for any two hyperparameters). In order to account for a possible scheduling of both $\alpha$ and $\lambda$, we introduce a scaling factor $\eta_t$ delivered by a user-defined procedure   $SetScheduleMultiplier(t)$. 

Now, let's turn to adaptive gradient algorithms like the popular optimizer Adam~\cite{kingma2014adam}, which scale gradients by their historic magnitudes. Intuitively, when Adam is run on a loss function $f$ plus L$_2$ regularization, weights that tend to have large gradients in $f$ do not get regularized as much as they would with decoupled weight decay, since the gradient of the regularizer gets scaled along with the gradient of $f$.
This leads to an inequivalence of L$_2$ and decoupled weight decay regularization for adaptive gradient algorithms:

\begin{prop}[Weight decay $\neq$ L$_2$ reg for adaptive gradients] Let $O$ denote an optimizer that has iterates $\bm{\theta}_{t+1} \leftarrow \bm{\theta}_t - \alpha \mathbf{M}_t \nabla f_t(\bm{\theta}_t)$ when run on batch loss function $f_t(\bm{\theta})$ \emph{without} weight decay, and 
$\bm{\theta}_{t+1} \leftarrow (1 - \lambda) \bm{\theta}_t - \alpha \mathbf{M}_t \nabla f_t(\bm{\theta}_t)$ when run on $f_t(\bm{\theta})$ \emph{with} weight decay, respectively, with $\mathbf{M}_t \neq k \mathbf{I}$ (where $k\in\mathbb{R}$). 
Then, for $O$ there exists no L$_2$ coefficient $\lambda'$ such that running $O$ on batch loss $f^{\text{reg}}_t(\bm{\theta}) = f_t(\bm{\theta}) + \frac{\lambda'}{2} \norm{\bm{\theta}}_2^2$ without weight decay is equivalent to running $O$ on $f_t(\bm{\theta})$ with decay $\lambda\in\mathbb{R}^+$. 
%For adaptive gradient algorithms with $\mathbf{M}_t \neq k \mathbf{I}$ (where $k\in\mathrm{R}$), there exists no L$_2$ regularizer $\lambda' \norm{\bm{\theta}}_2^2$ that is equivalent to the weight decay regularization defined in Equation \ref{eq:wdecay}.
\end{prop}

We decouple weight decay and loss-based gradient updates in Adam as shown in line 12 of Algorithm 2; this gives rise to our variant of Adam with decoupled weight decay (\textbf{AdamW}). %As we will demonstrate experimentally (in Section \ref{sec:exp_generalization}), AdamW generalizes much better than Adam with standard L$_2$ regularization.

%\note{FH: maybe here we could put/integrate some of your new text about explaining intuition as to why it may generalize better?}

%\note{FH: the non-speculative part of the ICML part added back.}
Having shown that L$_2$ regularization and weight decay regularization differ for adaptive gradient algorithms raises the question of how they differ and how to interpret their effects. 
Their equivalence for standard SGD remains very helpful for intuition: both mechanisms push weights closer to zero, at the same rate.
%However, for adaptive gradient algorithms they differ: with L$_2$ regularization, these adapt the summed gradient of the regularizer and the loss function, whereas with weight decay they only adapt the gradients of the loss function (with the weight decay step separated from the adaptive gradient mechanism).
However, for adaptive gradient algorithms they differ: with L$_2$  regularization, the sums of the gradient of the loss function and the gradient of the regularizer (i.e., the L$_2$  norm of the weights) are adapted, whereas with decoupled weight decay, only the gradients of the loss function are adapted (with the weight decay step separated from the adaptive gradient mechanism).
With L$_2$ regularization both types of gradients are normalized by their typical (summed) magnitudes, and therefore weights $x$ with large typical gradient magnitude $s$ are regularized by a smaller relative amount than other weights.
In contrast, decoupled weight decay regularizes all weights with the same rate $\lambda$, effectively regularizing weights $x$ with large $s$ more than standard L$_2$ regularization does.
%Penalizing weights with (typically) large gradients more than weights with (typically) small gradients makes intuitive sense because small variations in those weights have large effects, meaning that they make the solution more brittle. This also relates to the idea of sharp vs.\ flat local optima~\citep{keskar2016large}: penalizing weights with (typically) large gradients more heavily may drive the search away from areas where those gradients are large and thus lead to finding flatter optima with better generalization performance.
We demonstrate this formally for a simple special case of adaptive gradient algorithm with a fixed preconditioner:

%\begin{prop}[Fixed point of weight decay reg]
%Let $O$ denote an algorithm with the same characteristics as in Proposition 2, and using a fixed preconditioner matrix $\textbf{M}_t = \text{diag}(\vc{s})^{-1}$ ($\forall i. s_i>0$).
%Then, $O$ run on $f(\bm{\theta})$ with weight decay $\lambda$ and base learning rate $\alpha$ has a fixed point at the minimizer of the scale-adjusted regularized function
%\begin{equation}
%f_{\text{sreg}}(\bm{\theta}) = f(\bm{\theta}) + \frac{\lambda}{2\alpha} \norm{\bm{\theta} \odot{} \vc{s}}_2^2,
%\end{equation}
%where $\odot$ denotes element-wise multiplication.
%\end{prop}

\begin{prop}[Weight decay = scale-adjusted $L_2$ reg for adaptive gradient algorithm with fixed preconditioner]
Let $O$ denote an algorithm with the same characteristics as in Proposition 2, and using a fixed preconditioner matrix $\textbf{M}_t = \text{diag}(\vc{s})^{-1}$ (with $s_i>0$ for all $i$).
Then, $O$ with base learning rate $\alpha$ executes the same steps on batch loss functions $f_t(\bm{\theta})$ with weight decay $\lambda$ as it executes without weight decay on the scale-adjusted regularized batch loss 
\vspace*{-0.2cm}
\begin{equation}f_{t}^{\text{sreg}}(\bm{\theta}) = f_t(\bm{\theta}) + \frac{\lambda'}{2\alpha} \norm{\bm{\theta} \odot{} \sqrt{\vc{s}}}_2^2,\vspace*{-0.1cm}
\end{equation}
where $\odot$ and $\sqrt{\cdot}$ denote element-wise multiplication and square root, respectively, and $\lambda' = \frac{\lambda}{\alpha}$.
\end{prop}

We note that this proposition does \emph{not} directly apply to practical adaptive gradient algorithms, since these change the preconditioner matrix at every step. Nevertheless, it can still provide intuition about the equivalent loss function being optimized in each step: parameters $\theta_i$ with a large inverse preconditioner $s_i$ (which in practice would be caused by historically large gradients in dimension $i$) are regularized relatively more than they would be with L$_2$ regularization; specifically, the regularization is proportional to $\sqrt{s_i}$.

\section{Justification of Decoupled Weight Decay via a View of Adaptive Gradient Methods as Bayesian Filtering}
\label{sec:justification}

We now discuss a justification of decoupled weight decay in the framework of Bayesian filtering for a unified theory of adaptive gradient algorithms due to \citet{aitchison18}.
After we posted a preliminary version of our current paper on arXiv, Aitchison noted that his theory ``gives us a theoretical framework in which we can understand the superiority of this weight decay over $L_2$ regularization, because it is weight decay, rather than $L_2$ regularization that emerges through the straightforward application of Bayesian filtering.''\citep{aitchison18}.
While full credit for this theory goes to Aitchison, we summarize it here to shed some light on why weight decay may be favored over $L_2$ regularization.

\citet{aitchison18} views stochastic optimization of $n$ parameters $\theta_1, \dots, \theta_n$ as a Bayesian filtering problem with the goal of inferring a distribution over the optimal values of each of the parameters $\theta_i$ given the current values of the other parameters $\bm{\theta}_{-i}(t)$ at time step $t$. When the other parameters do not change this is an optimization problem, but when they do change it becomes one of ``tracking'' the optimizer using Bayesian filtering as follows. One is given a probability distribution $P(\bm{\theta}_{t} \mid \bm{y_{1:t}})$ of the optimizer at time step $t$ that takes into account the data $\bm{y_{1:t}}$ from the first $t$ mini batches, a state transition prior $P(\bm{\theta}_{t+1} \mid \bm{\theta}_t)$ reflecting a (small) data-independent change in this distribution from one step to the next, and a likelihood $P(\bm{y}_{t+1} \mid \bm{\theta}_{t+1})$ derived from the mini batch at step $t+1$. The posterior distribution $P(\bm{\theta}_{t+1} \mid \bm{y_{1:t+1}})$ of the optimizer at time step $t+1$ can then be computed (as usual in Bayesian filtering) by marginalizing over $\bm{\theta}_{t}$ to obtain the one-step ahead predictions $P(\bm{\theta}_{t+1} \mid \bm{y_{1:t}})$ and then applying Bayes' rule to incorporate the likelihood $P(\bm{y}_{t+1} \mid \bm{\theta}_{t+1})$. \citet{aitchison18} assumes a Gaussian state transition distribution $P(\bm{\theta}_{t+1} \mid \bm{\theta}_t)$ and an approximate conjugate likelihood $P(\bm{y}_{t+1} \mid \bm{\theta}_{t+1})$, leading to the following closed-form update of the filtering distribution's mean:
\begin{equation}
\bm{\mu}_{post} = \bm{\mu}_{prior} + \bm{\Sigma}_{post} \times \bm{g},
\end{equation}
where $\bm{g}$ is the gradient of the log likelihood of the mini batch at time $t$.
This result implies a preconditioner of the gradients that is given by the posterior uncertainty $\bm{\Sigma}_{post}$ of the filtering distribution: updates are larger for parameters we are more uncertain about and smaller for parameters we are more certain about. 
\citet{aitchison18} goes on to show that popular adaptive gradient methods, such as Adam and RMSprop, as well as Kronecker-factorized methods are special cases of this framework.

Decoupled weight decay very naturally fits into this unified framework as part of the state-transition distribution: \citet{aitchison18} assumes a slow change of the optimizer according to the following Gaussian:
\begin{equation}
\label{eq:aitchison}P(\bm{\theta}_{t+1} \mid \bm{\theta}_t) = \mathcal{N}((\bm{I}-\bm{A}) \bm{\theta}_t, \bm{Q}),
\end{equation}

where $\bm{Q}$ is the covariance of Gaussian perturbations of the weights, and $\bm{A}$ is a regularizer to avoid values growing unboundedly over time. When instantiated as $\bm{A} = \lambda\times \bm{I}$, this regularizer $\bm{A}$ plays exactly the role of decoupled weight decay as described in Equation \ref{eq:wdecay}, since this leads to multiplying the current mean estimate $\bm{\theta}_t$ by $(1-\lambda)$ at each step. Notably, this regularization is also directly applied to the prior and does not depend on the uncertainty in each of the parameters (which would be required for $L_2$ regularization).

\begin{figure*}[t]%
\begin{center}
  \includegraphics[width=0.27\textwidth]{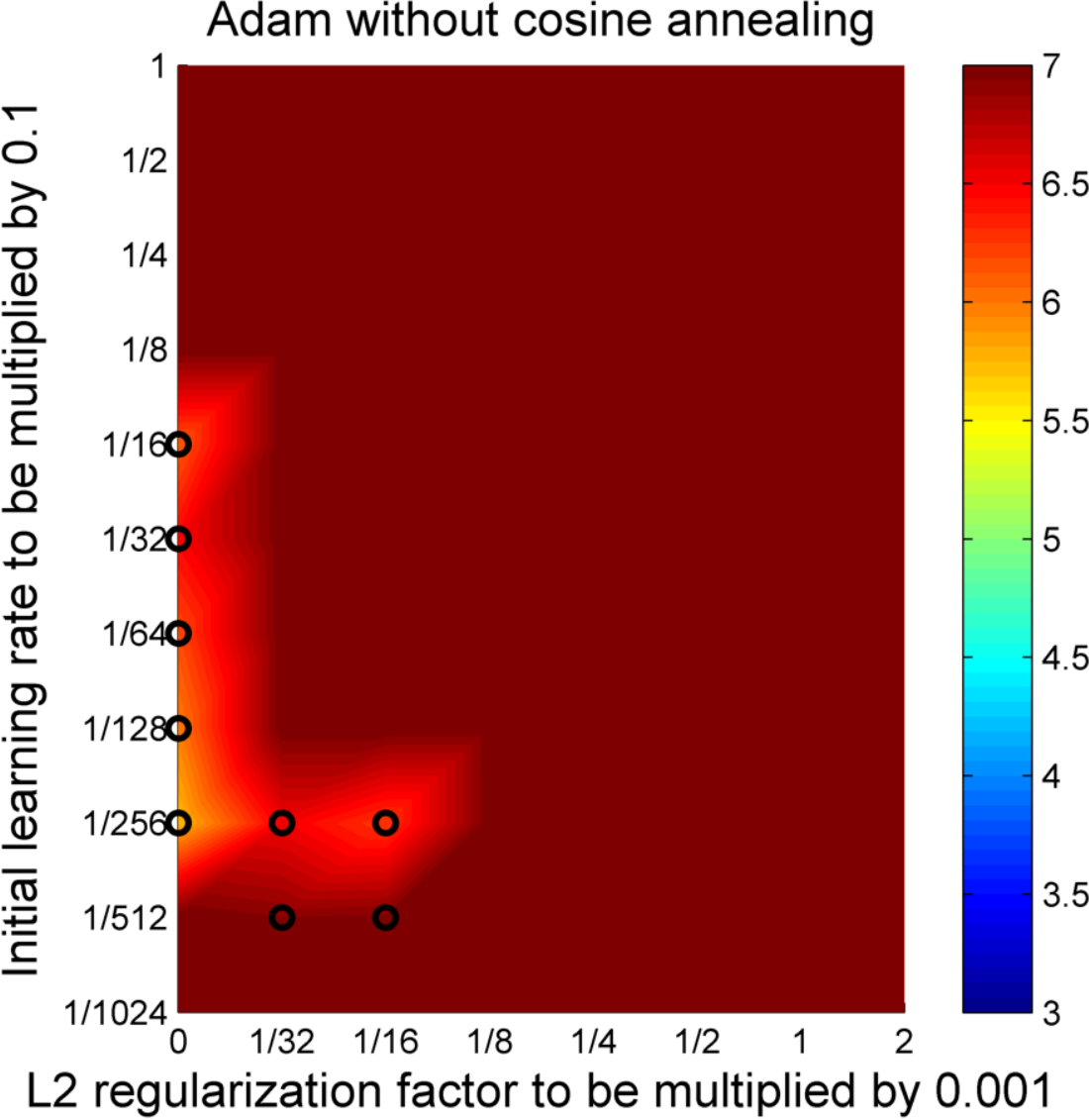}~
    \includegraphics[width=0.27\textwidth]{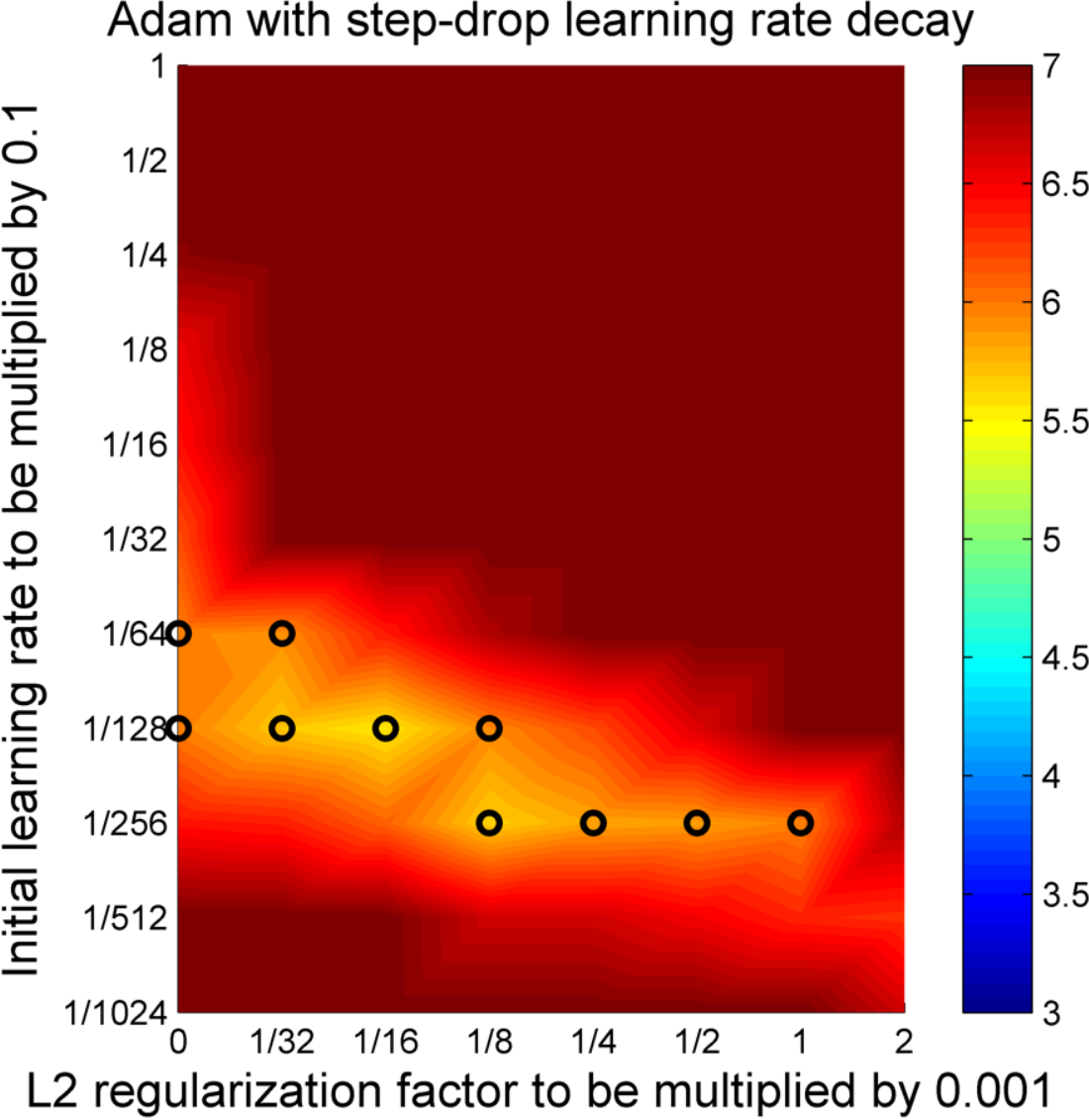}~
  \includegraphics[width=0.27\textwidth]{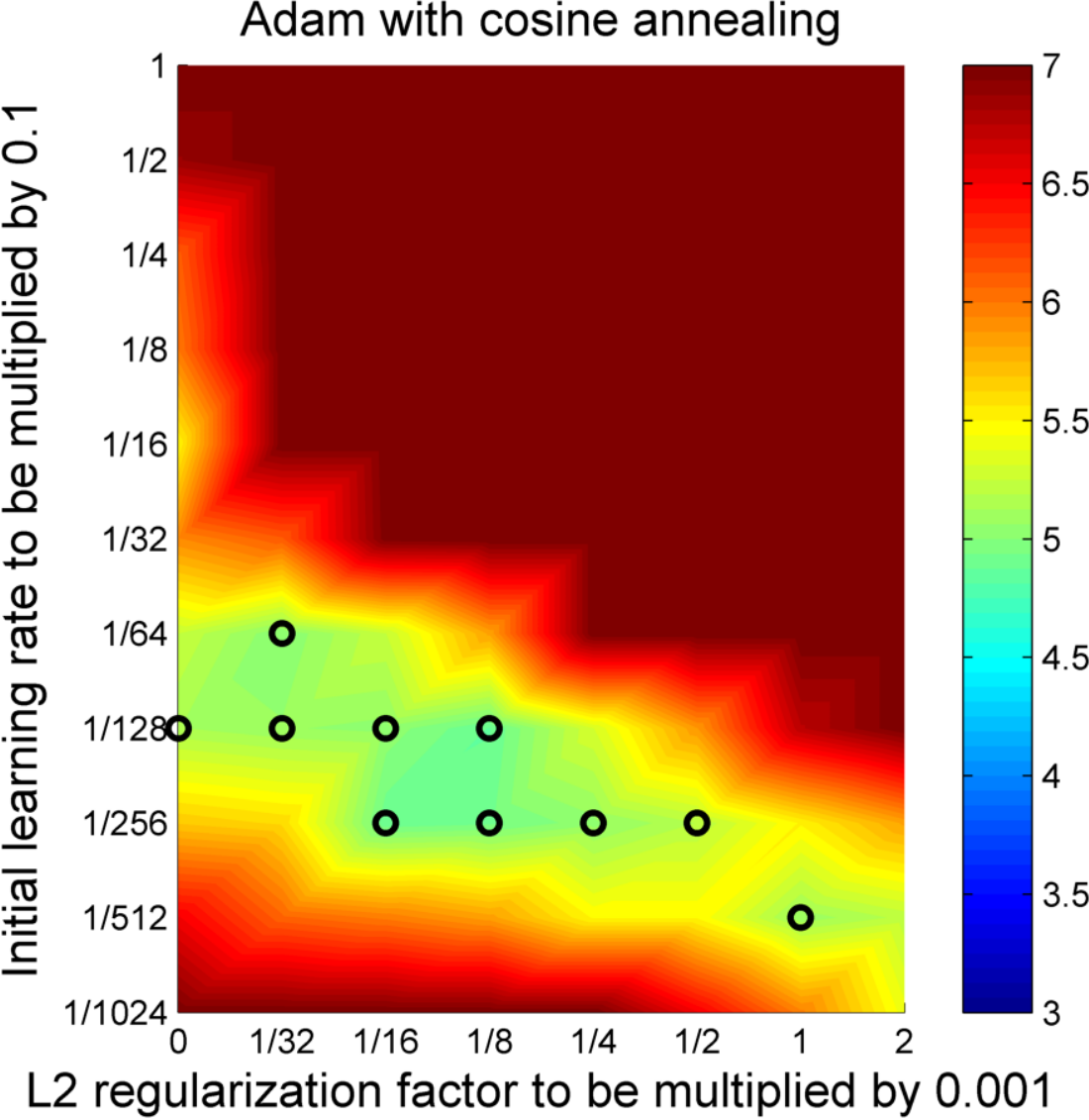}~\\
  \includegraphics[width=0.27\textwidth]{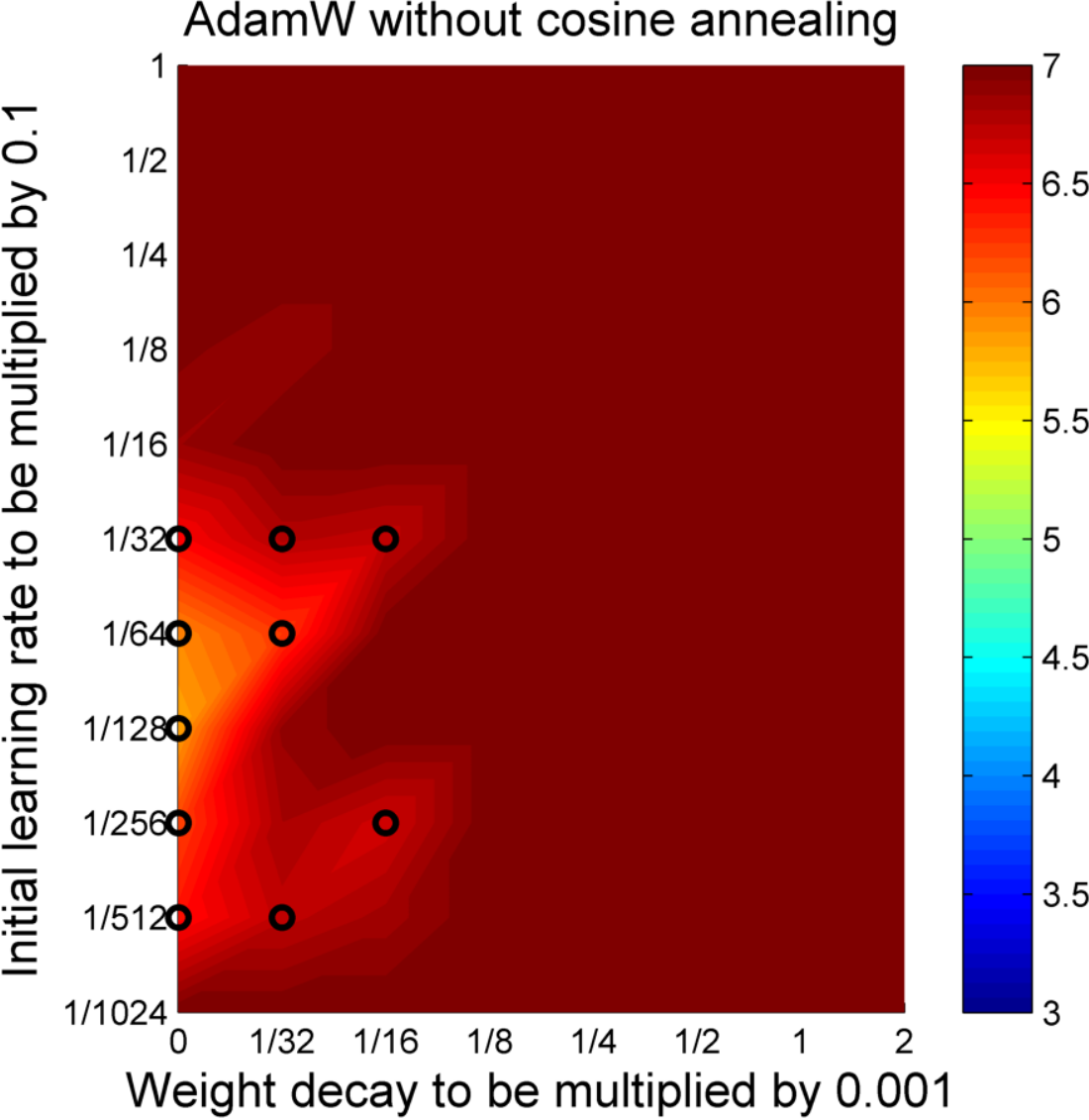}~
  \includegraphics[width=0.27\textwidth]{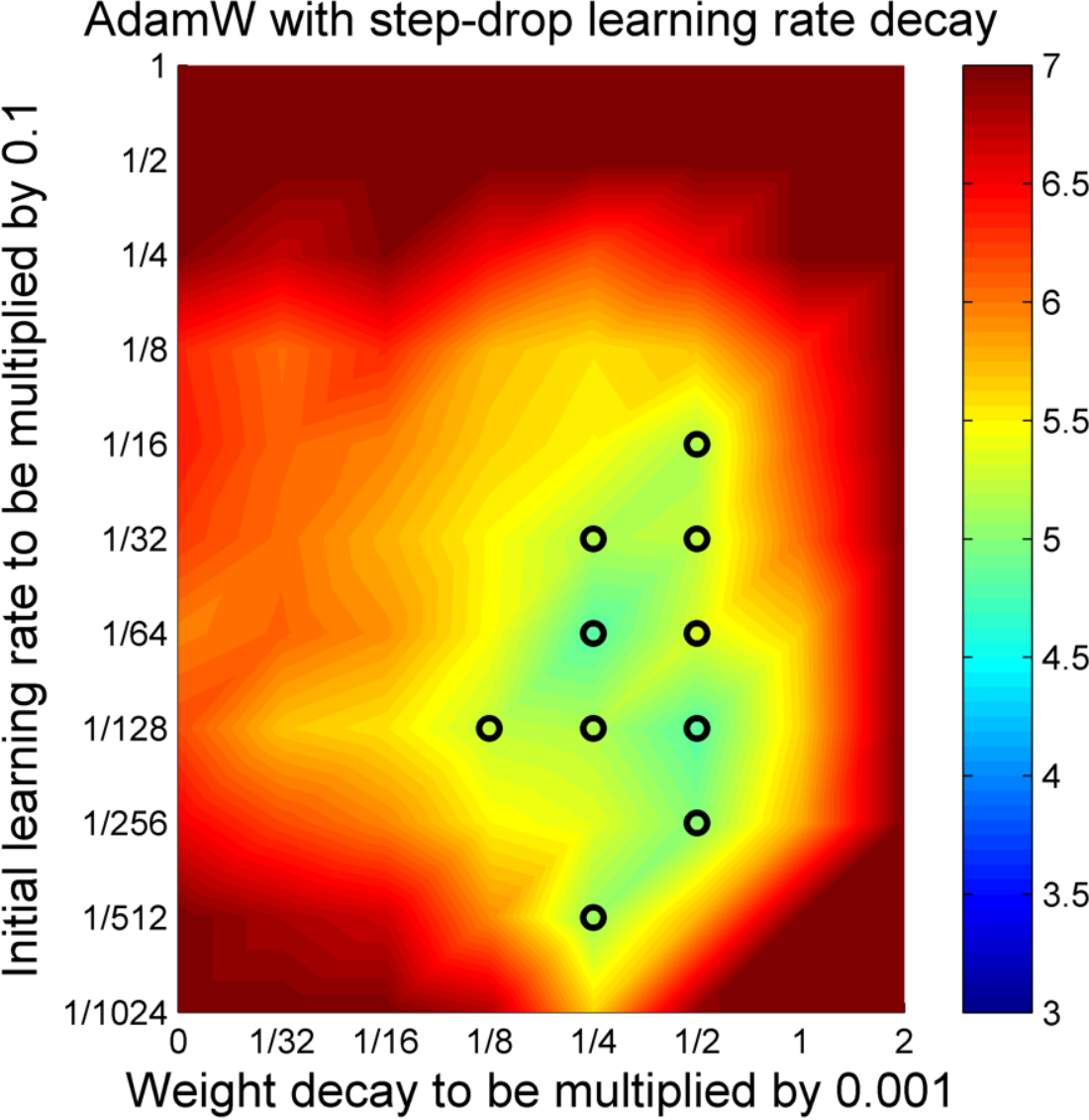}~
  \includegraphics[width=0.27\textwidth]{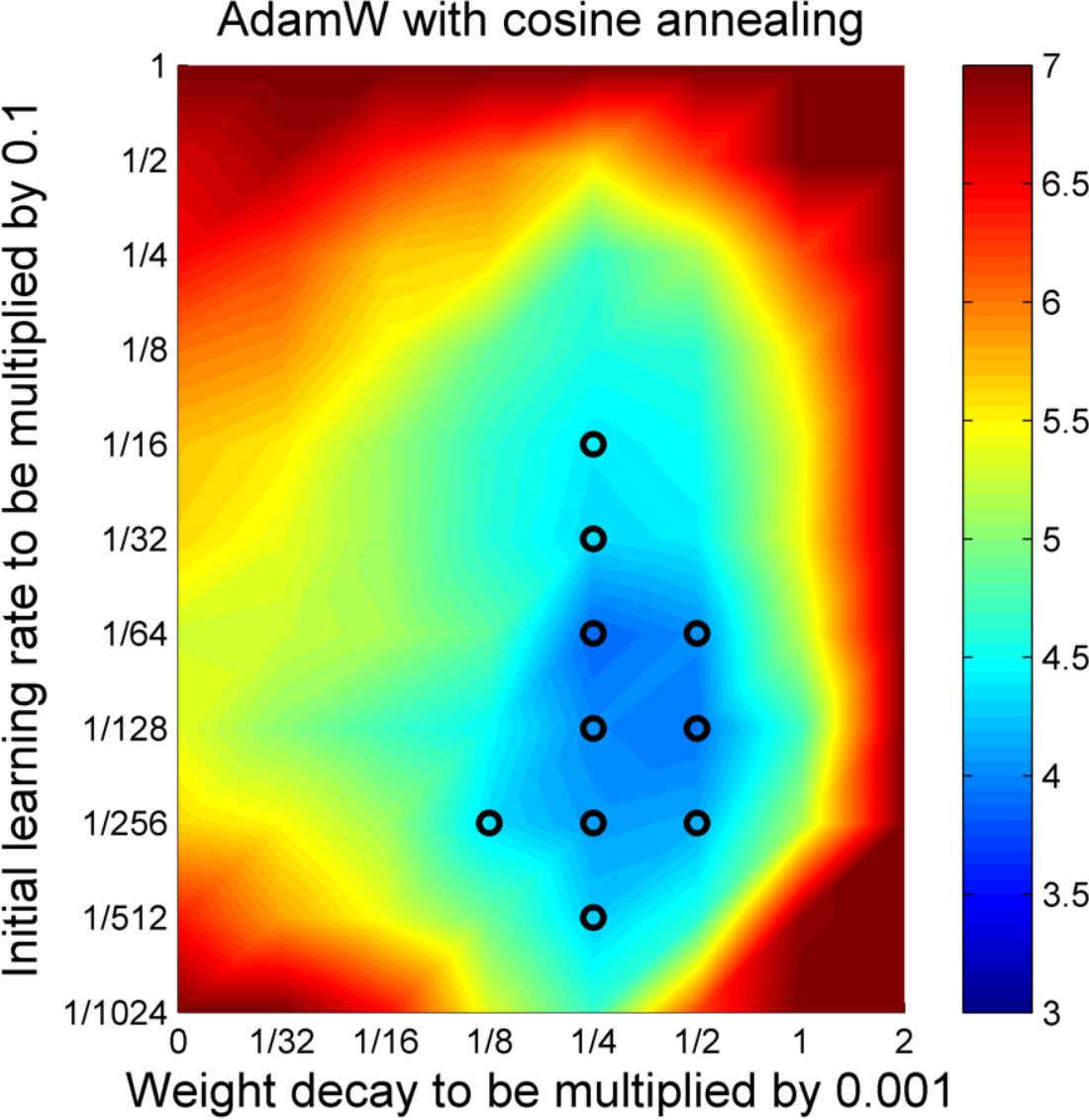}\\
  %\vspace*{-0.5cm}
\caption{\label{fig:adam_with_without_cosine_stepdrop} Adam performs better with decoupled weight decay (bottom row, AdamW) than with $L_2$ regularization (top row, Adam). We show the final test error of a 26 2x64d ResNet on CIFAR-10 after 100 epochs of training with fixed learning rate (left column), step-drop learning rate (with drops at epoch indexes 30, 60 and 80, middle column) and cosine annealing (right column). AdamW leads to a more separable hyperparameter search space, especially when a learning rate schedule, such as step-drop and cosine annealing is applied. Cosine annealing yields clearly superior results. 
%Therefore, we schedule the learning rate with cosine annealing for all methods given in the paper.
}
\end{center}
\end{figure*}

%\citet{aitchison18} also points out another issue of $L_2$ regularization when viewed from a Bayesian perspective that is fixed by decoupled weight decay. The usual interpretation of $L_2$ regularization is that it corresponds to gradient descent on an objective function that is regularized by an $L_2$ loss, and the standard Bayesian interpretation is that this regularizer is the log probability of an IID Gaussian prior distribution over the weights. However, in Bayesian statistics, priors get overwhelmed when enough data is available. The degree of weight decay induced by $L_2$ regularization would therefore scale as the inverse of the number of mini-batches and thus vanish for long training runs, but empirically, positive weight decay values remain useful even for very large datasets. In contrast, under the interpretation of Equation \ref{eq:aitchison}, which is compatible with decoupled weight decay, weight decay remains active even in the limit of an infinite number of weight updates.

\section{Experimental Validation}

%We experimentally validate the proposed way to decouple the weight decay and the gradient-based update in SGD with momentum and Adam. The results given in section 5.1 will demonstrate that the decoupling is possible and simplifies the setting of hypeparameters. 
%In section 5.2, we validate AdamR designed to improve any-time performance by deliving good solutions earlier. 

%\begin{figure}[tbp]%
%\begin{center}
%	\includegraphics[width=0.24\textwidth]{fig1_ADAMwithoutcosine.pdf}~
%  \includegraphics[width=0.24\textwidth]{fig1_ADAMwithcosine.pdf}
%  \vspace*{-0.5cm}
%\caption{\label{fig:adam_with_without_cosine} We show the final test error of a 26 2x64d ResNet on CIFAR-10 after 100 epochs of standard Adam with L2 regularization, with fixed learning rate (left) and with cosine annealing (right). Cosine annealing yields clearly superior results. 
%See SuppFigure 2 %%\ref{fig:adam_with_without_cosine_stepdrop} 
%(in the supplementary material) for a comparison with a step-drop learning rate schedule.
%%Therefore, we schedule the learning rate with cosine annealing for all methods given in the paper.
%}
%\end{center}
%\end{figure}

We now evaluate the performance of decoupled weight decay under various training budgets and learning rate schedules.
Our experimental setup follows that of \citet{gastaldi2017shake}, who proposed, in addition to L$_2$ regularization, to apply the new Shake-Shake regularization to a 3-branch residual DNN that allowed %. \citet{gastaldi2017shake} showed that this regularization allowed 
to achieve new state-of-the-art results of 2.86\% on the CIFAR-10 dataset~\citep{krizhevsky2009learning}.
%and of 15.85\% on CIFAR-100. 
We used the same model/source code based on fb.resnet.torch \footnote{https://github.com/xgastaldi/shake-shake}. 
We always used a batch size of 128 and applied 
%for 1800 epochs with . %with the learning rate scheduled by Eq. (\ref{eq:t}). 
the regular data augmentation procedure for the CIFAR datasets. 
The base networks are a 26 2x64d ResNet (i.e. the network has a depth of 26, 2 residual branches and the first residual block has a width of 64) and a 26 2x96d ResNet with 11.6M and 25.6M parameters, respectively. 
For a detailed description of the network and the Shake-Shake method, we refer the interested reader to \citet{gastaldi2017shake}. We also perform experiments on the ImageNet32x32 dataset \citep{chrabaszcz2017downsampled}, a downsampled version of the original ImageNet dataset with 1.2 million  32$\times$32 pixels images.

\subsection{Evaluating Decoupled Weight Decay With Different Learning Rate Schedules}

In our first experiment, we compare Adam with $L_2$ regularization to Adam with decoupled weight decay (AdamW), using three different learning rate schedules: a fixed learning rate, a drop-step schedule, and a cosine annealing schedule~\citep{loshchilov2016sgdr}.
Since Adam already adapts its parameterwise learning rates it is not as common to use a learning rate multiplier schedule with it as it is with SGD, but as our results show such schedules can substantially improve Adam's performance, and we advocate not to overlook their use for adaptive gradient algorithms. 

For each learning rate schedule and weight decay variant, we trained a 2x64d ResNet for 100 epochs, using different settings of the initial learning rate $\alpha$ and the weight decay factor $\lambda$. 
Figure \ref{fig:adam_with_without_cosine_stepdrop} shows that decoupled weight decay outperforms $L_2$ regularization for all 
learning rate schedules, with larger differences for better learning rate schedules. We also note that decoupled weight decay leads to a more separable hyperparameter search space, especially when a learning rate schedule, such as step-drop and cosine annealing is applied.
The figure also shows that cosine annealing clearly outperforms the other learning rate schedules; we thus used cosine annealing for the remainder of the experiments.

\begin{figure*}[t]
\begin{center}
    \includegraphics[width=0.35\textwidth]{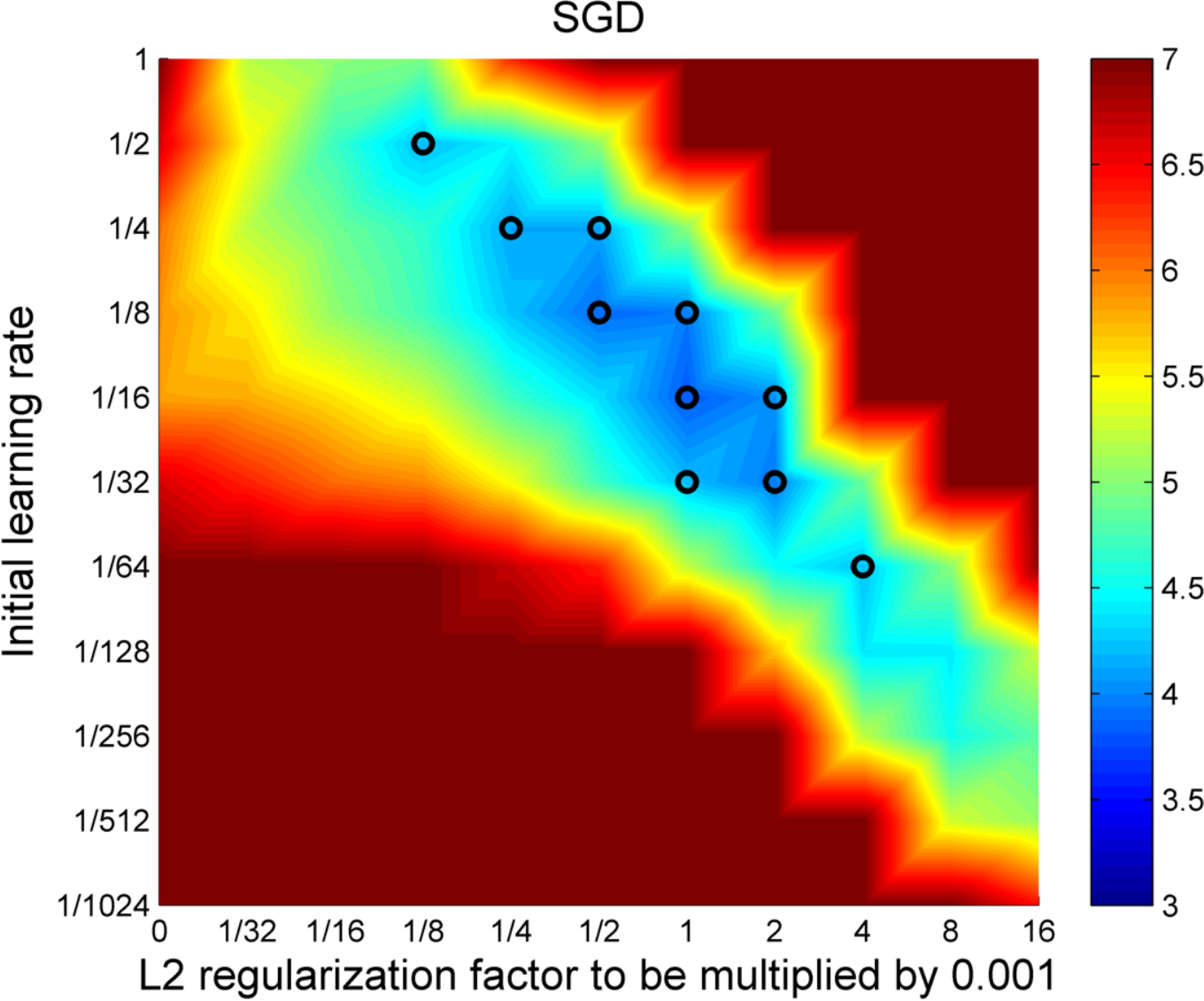} $\;\;$~~~~~~~
  \includegraphics[width=0.35\textwidth]{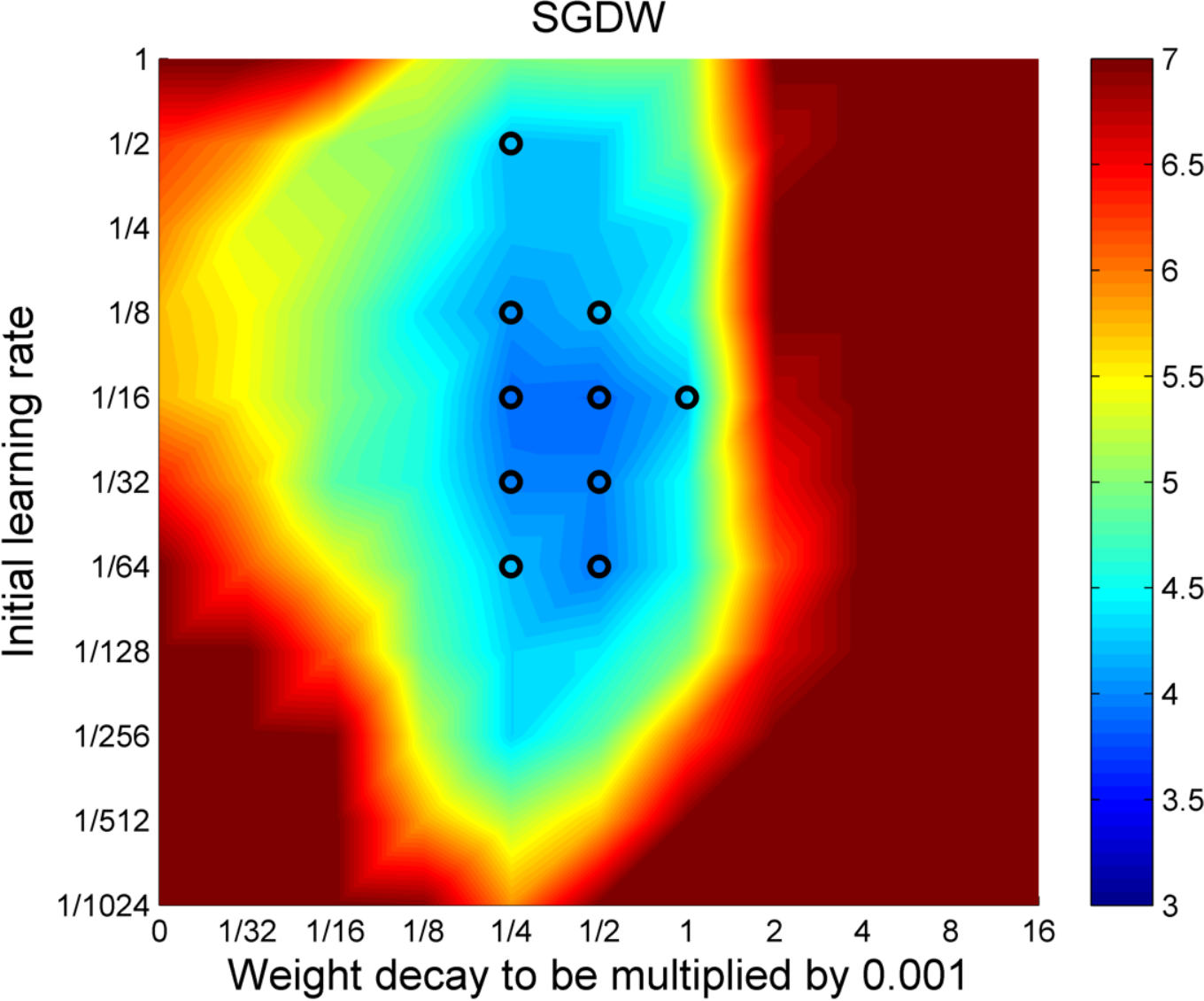}\\
	$\;\;$\\
	\includegraphics[width=0.35\textwidth]{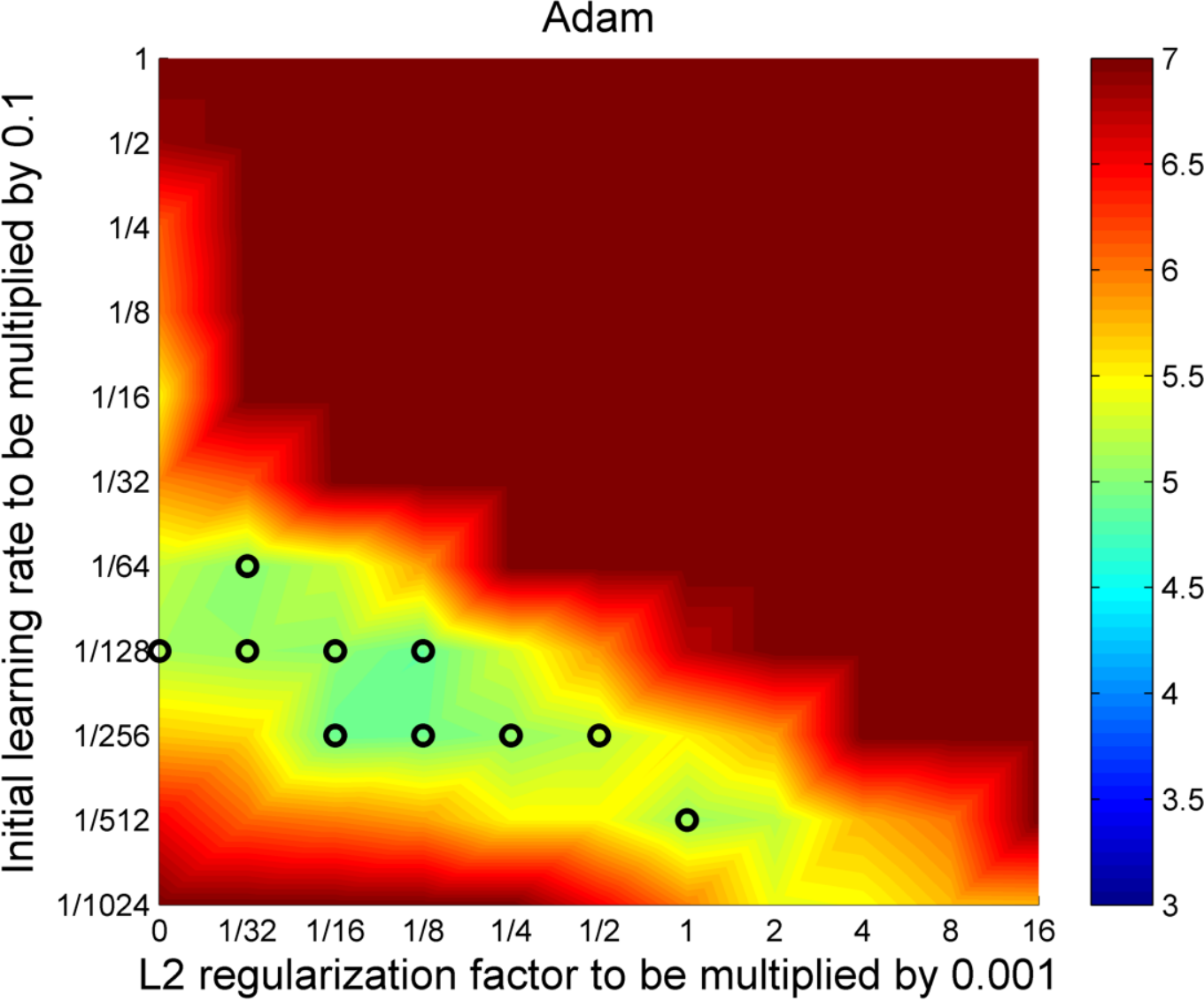} $\;\;$~~~~~~~
  \includegraphics[width=0.35\textwidth]{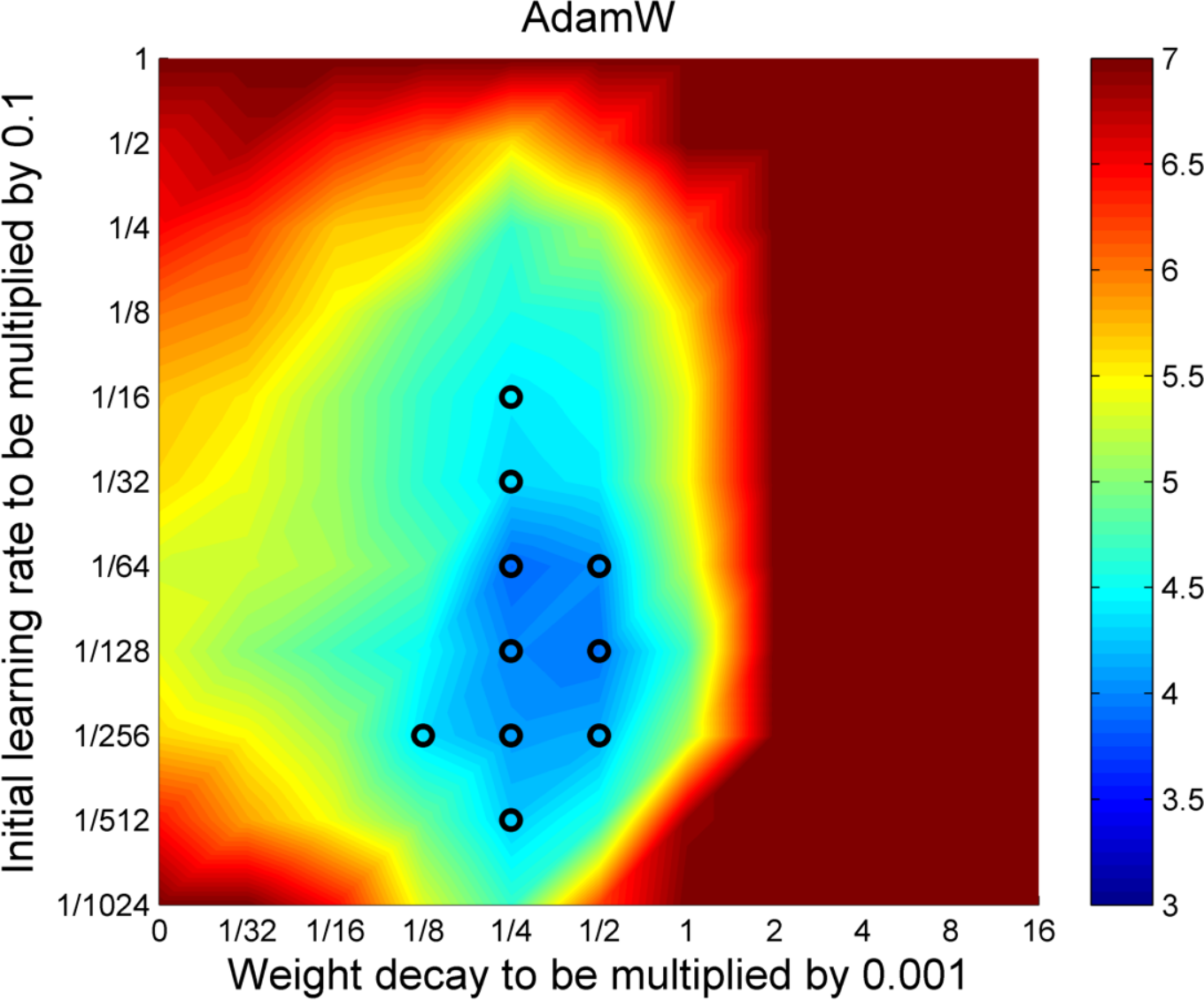}
\caption{\label{fig2_hyp100epochs} The Top-1 test error of a 26 2x64d ResNet on CIFAR-10 measured after 100 epochs. The proposed SGDW and AdamW (right column)  have a more separable hyperparameter space.}
\vspace*{-0.25cm}
\end{center}
\end{figure*}

\begin{figure*}[t]%
\begin{center}
	\includegraphics[width=0.35\textwidth]{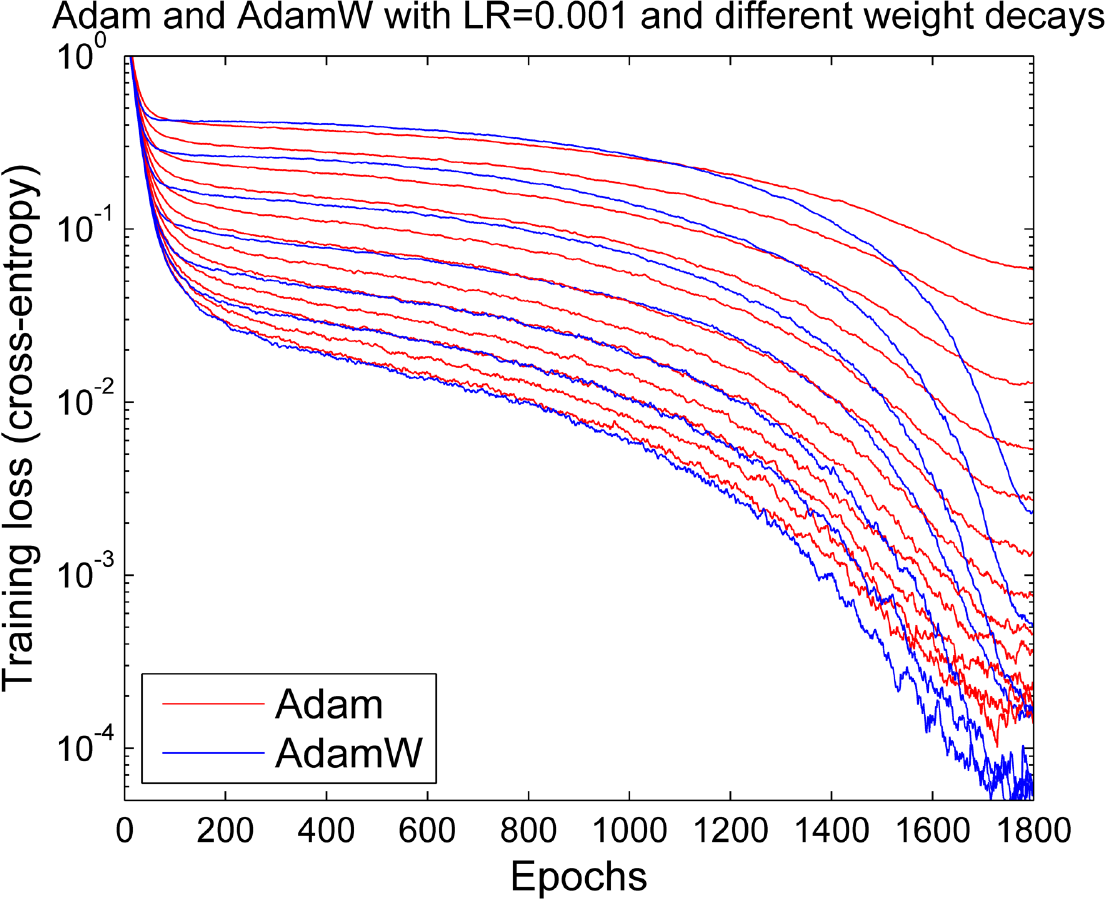} $\;\;$~~~~~~~~~~~
  \includegraphics[width=0.35\textwidth]{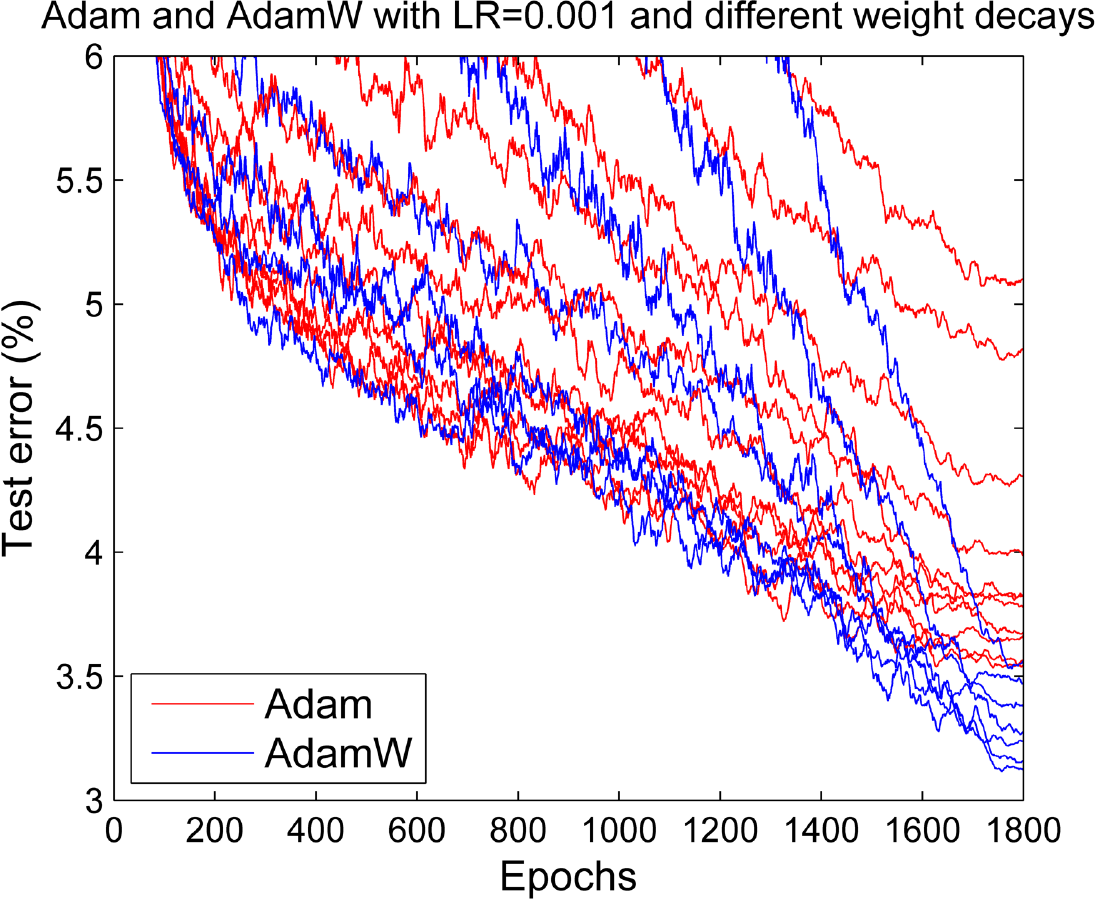}\\
	$\;\;$\\
	\includegraphics[width=0.35\textwidth]{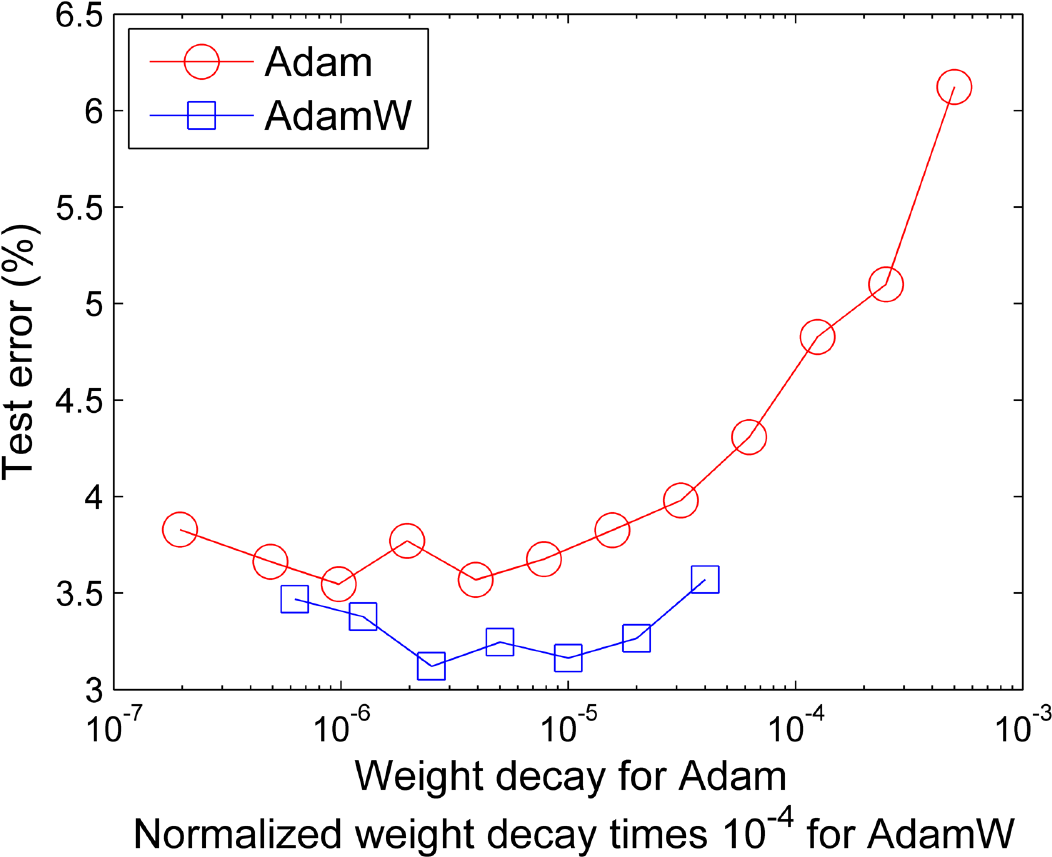} $\;\;$~~~~~~~~~~~
    \includegraphics[width=0.35\textwidth]{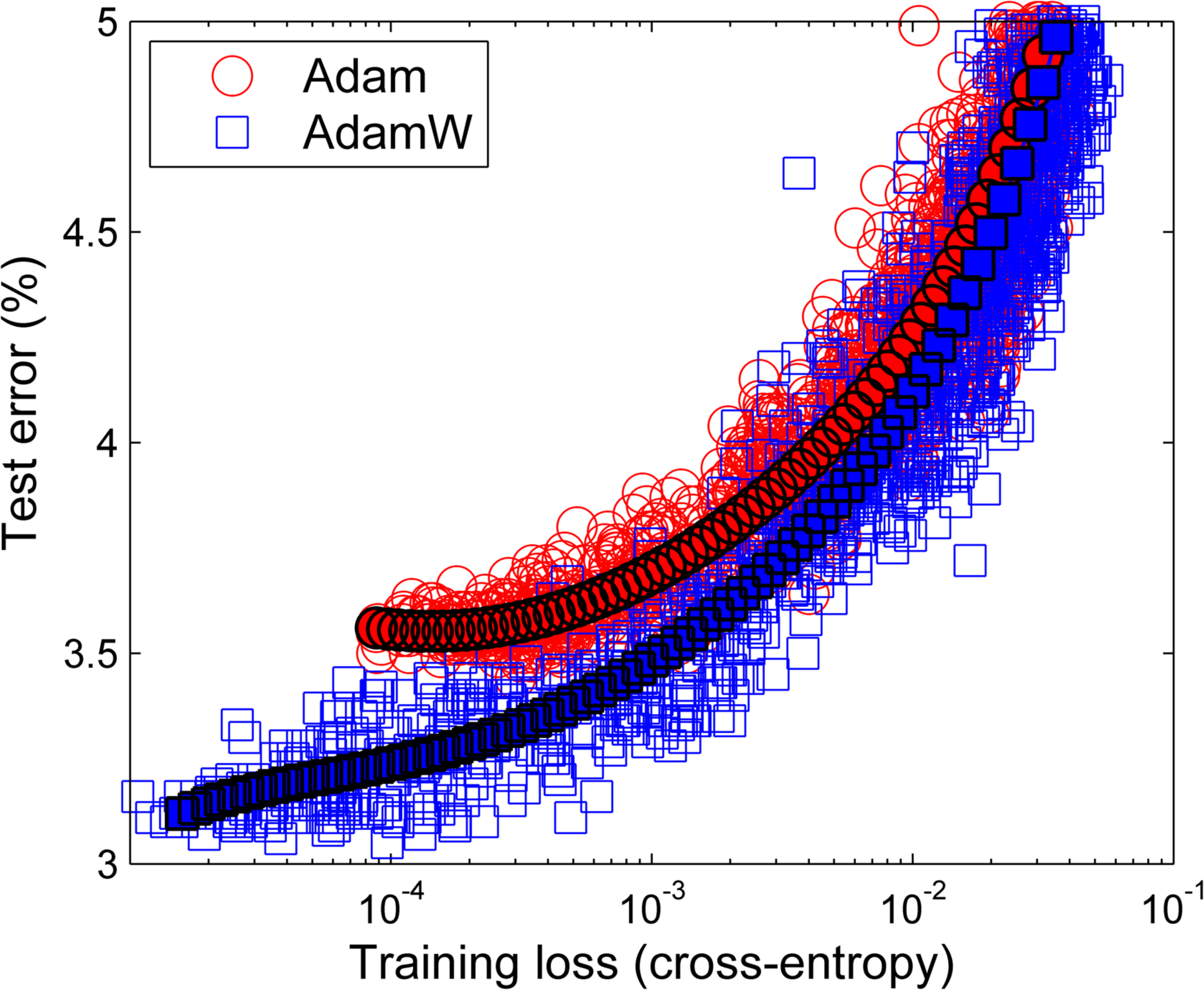}
\vspace*{-0.2cm}\caption{\label{fig1800} Learning curves (top row) and generalization results (bottom row) obtained by a 26 2x96d ResNet trained with Adam and AdamW on CIFAR-10. 
See text for details. SuppFigure 4 in the Appendix shows the same qualitative results for ImageNet32x32.}\vspace*{-0.25cm}
\end{center}
\end{figure*}

\subsection{Decoupling the Weight Decay and Initial Learning Rate Parameters}

In order to verify our hypothesis about the coupling of $\alpha$ and  $\lambda$, in Figure \ref{fig2_hyp100epochs} we compare the performance of L$_2$ regularization vs.\ decoupled weight decay in SGD (SGD vs.\ SGDW, top row) and in Adam (Adam vs.\ AdamW, bottom row). In SGD (Figure \ref{fig2_hyp100epochs}, top left), L$_2$ regularization is not decoupled from the learning rate (the common way as described in Algorithm 1), and the figure clearly shows that the basin of best hyperparameter settings (depicted by color and top-10 hyperparameter settings by black circles) is not aligned with the x-axis or y-axis but lies on the diagonal. This suggests that the two hyperparameters are interdependent and need to be changed simultaneously, while only changing one of them might substantially worsen results. Consider, e.g., the setting at the top left black circle ($\alpha=1/2$, $\lambda=1/8*0.001$); only changing either $\alpha$ or $\lambda$ by itself would worsen results, while changing both of them could still yield clear improvements. We note that this coupling of initial learning rate and L$_2$ regularization factor might have contributed to SGD's reputation of being very sensitive to its hyperparameter settings. 
%The latter might complicate the procedure of  hyperparameter selection especially when performed by some sort of one-dimensional / coordinate search as sometimes done when the number of function evaluation is extremely small.

In contrast, the results for SGD with decoupled weight decay (SGDW) in Figure \ref{fig2_hyp100epochs} (top right) show that weight decay and initial learning rate are decoupled. The proposed approach renders the two hyperparameters more separable: even if the learning rate is not well tuned yet (e.g., consider the value of 1/1024 in Figure \ref{fig2_hyp100epochs}, top right), leaving it fixed and only optimizing the weight decay factor would yield a good value (of 1/4*0.001). This is not the case for SGD with L$_2$ regularization (see Figure \ref{fig2_hyp100epochs}, top left).

The results for Adam with L$_2$ regularization are given in Figure \ref{fig2_hyp100epochs} (bottom left). Adam's best hyperparameter settings performed clearly worse than SGD's best ones (compare Figure \ref{fig2_hyp100epochs}, top left). While both methods used L$_2$ regularization,  Adam did not benefit from it at all: its best results obtained for non-zero L$_2$ regularization factors were comparable to the best ones obtained without the L$_2$ regularization, i.e., when $\lambda=0$.   
Similarly to the original SGD, the shape of the hyperparameter landscape suggests that the two hyperparameters are coupled. 

In contrast, the results for our new variant of Adam with decoupled weight decay (AdamW) in Figure \ref{fig2_hyp100epochs} (bottom right) show that AdamW largely decouples weight decay and learning rate. The results for the best hyperparameter settings were substantially better than the best ones of Adam with L$_2$ regularization and rivaled those of SGD and SGDW. 

In summary, the results in Figure \ref{fig2_hyp100epochs} support our hypothesis that the weight decay and learning rate hyperparameters can be decoupled, and that this in turn simplifies the problem of hyperparameter tuning in SGD and improves Adam's performance to be competitive w.r.t.\ SGD with momentum.

\subsection{Better Generalization of AdamW}\label{sec:exp_generalization}

While the previous experiment 
%results given in Figure \ref{fig2_hyp100epochs} 
suggested that the basin of optimal hyperparameters of AdamW is broader and deeper than the one of Adam, we next investigated the results for much longer runs of 1800 epochs to compare the generalization capabilities of AdamW and Adam.

We fixed the initial learning rate to 0.001 which represents both the default learning rate for Adam and the one which showed reasonably good results in our experiments. 
Figure \ref{fig1800} shows the results for 12 settings of the L$_2$ regularization of Adam and 7 settings of the normalized weight decay of AdamW (the normalized weight decay represents a rescaling formally defined in Appendix \ref{sec:normw}; it amounts to a multiplicative factor which depends on the number of batch passes). 
Interestingly, while the dynamics of the learning curves of Adam and AdamW often coincided for the first half of the training run, AdamW often led to lower training loss and test errors (see Figure \ref{fig1800} top left and top right, respectively). 
Importantly, the use of L$_2$ weight decay in Adam did not yield as good results as decoupled weight decay in AdamW (see also Figure \ref{fig1800}, bottom left).
%, because the weight decay of Adam does not decay weights exponentially. 
Next, we investigated whether AdamW's better results were only due to better convergence or due to better generalization. 
\emph{The results in Figure \ref{fig1800} (bottom right) for the best settings of Adam and AdamW suggest that AdamW did not only yield better training loss but 
also yielded better generalization performance for similar training loss values}.
%by delivering better test error values 
The results on ImageNet32x32 (see SuppFigure 4
%\ref{fig64ImageNet}
in the Appendix) yield the same conclusion of substantially improved generalization performance.

\begin{figure*}[!t]
\begin{center}
	\includegraphics[width=0.4\textwidth]{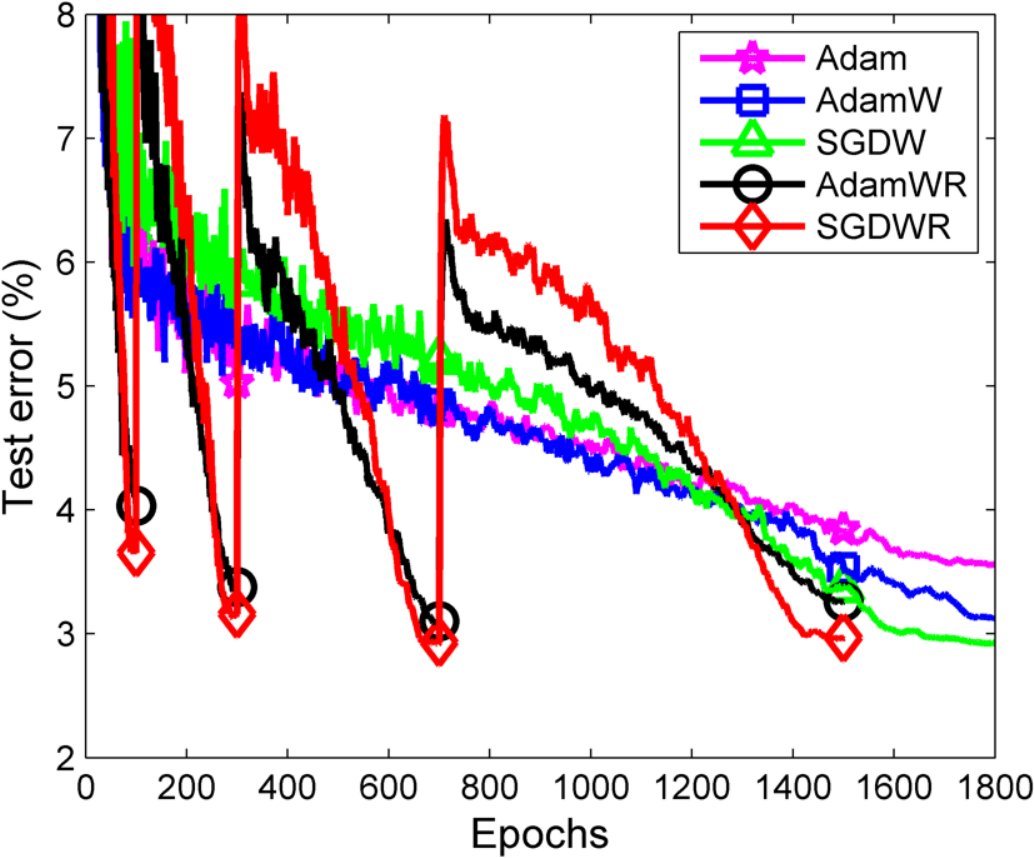} $\;\;$~~~~~~~~~~~
  \includegraphics[width=0.4\textwidth]{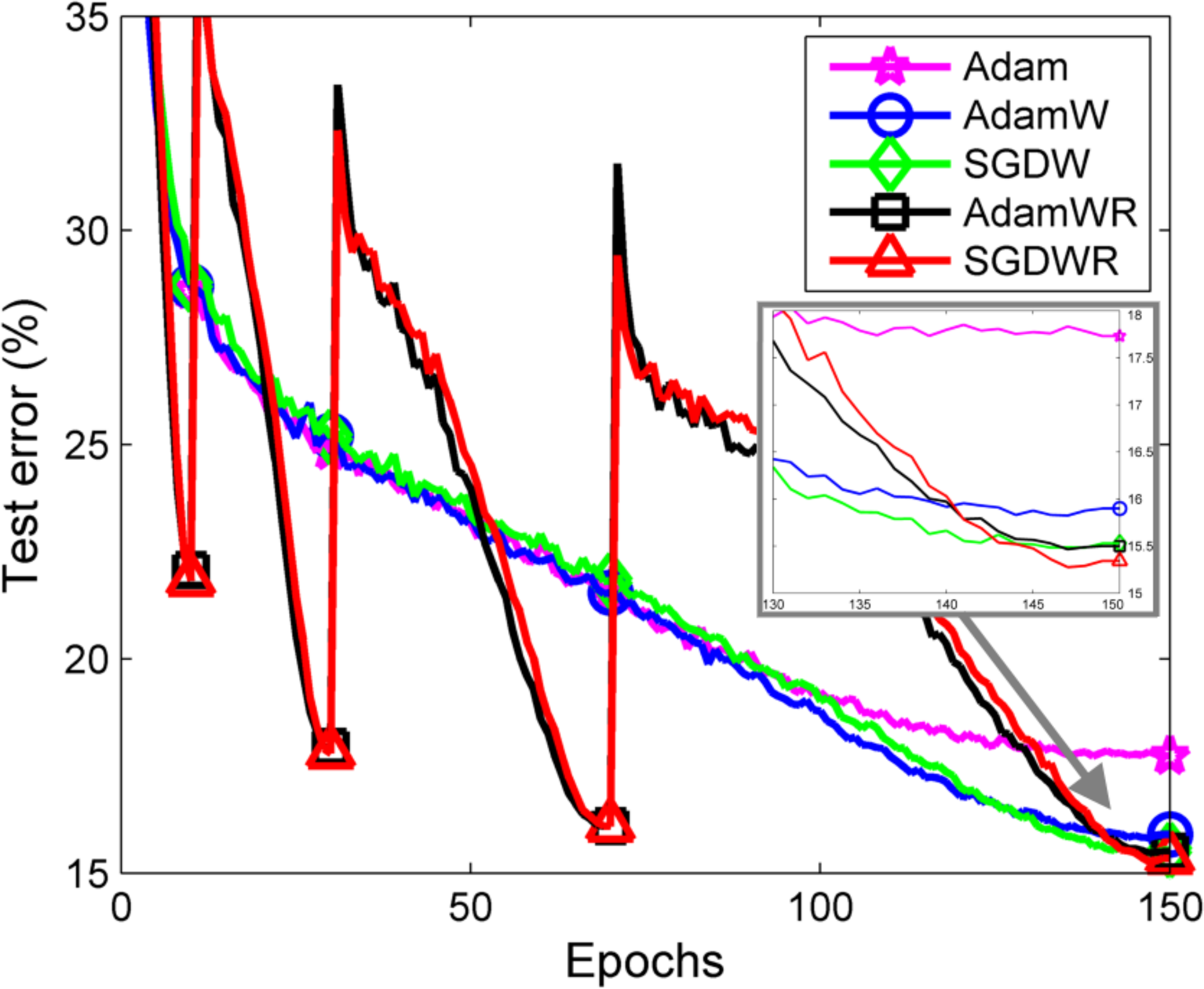}
\vspace*{-0.25cm}
\caption{\label{figrestarts} Top-1 test error on CIFAR-10 (left) and Top-5 test error on ImageNet32x32 (right). For a better resolution and with training loss curves, see SuppFigure \ref{figtrainingCIFAR} and SuppFigure \ref{figtrainingImagenet32} in the supplementary material.}% with and without snapshot-ensembles.}
\vspace*{-0.25cm}
\end{center}
\end{figure*}

\subsection{AdamWR with Warm Restarts for Better Anytime Performance}

%FH: reworded in the non-anonymous version
%following \cite{loshchilov2016sgdr}, who proposed to improve anytime performance of SGD by performing warm restarts,  we investigated AdamWR as an extension of AdamW (using normalized weight decay to avoid the need for a different weight decay factor for restarts with longer annealing schedules). 
In order to improve the anytime performance of SGDW and AdamW we extended them with the warm restarts we introduced in \citet{loshchilov2016sgdr}, to obtain SGDWR and AdamWR, respectively (see Section \ref{sec:adamwr} in the Appendix). %We investigated the strong anytime performance AdamWR obtains from warm restarts. %(using normalized weight decay to avoid the need for a different weight decay factor for restarts with longer annealing schedules).
As Figure \ref{figrestarts} shows, AdamWR greatly sped up AdamW on CIFAR-10 and ImageNet32x32, up to a factor of 10 (see the results at the first restart).
For the default learning rate of 0.001, \emph{AdamW achieved 15\% relative improvement in test error compared to Adam both on CIFAR-10} (also see SuppFigure  \ref{figtrainingCIFAR}) \emph{and ImageNet32x32} %(also see SuppFigure 5).
(also see SuppFigure \ref{figtrainingImagenet32}). 

\emph{AdamWR achieved the same improved results but with a much better anytime performance.} 
These improvements closed most of the gap between Adam and SGDWR on CIFAR-10 
%\note{FH: Can we put this into perspectice? How bad would the original Adam have been? 
%Optimal would be a curve for original Adam that is far worse than AdamWR. I know to be fair you'd like to do a full hyperparameter optimization for Adam, but just running its defaults would be a good start. And I think it is important to at least give some ballpark number to put this great result in perspective (since Adam is usually far worse here). Otherwise, reviewers may see this as a negative result.}
and yielded comparable performance on ImageNet32x32. 
%
%Additionally, one can build Snapshot-Ensembles \citep{SnapshotICLR2017} based on models obtained before each warm-restart. The final ensemble does not require any additional computational costs at training. \citep{SnapshotICLR2017, loshchilov2016sgdr} showed that the quality of snapshot-based ensembles is as good as the one obtained from independent runs with comparable training time costs. The presented results without ensembles match the state-of-the-art results of \citep{gastaldi2017shake} for CIFAR-10 and \cite{ImageNet32x32} for ImageNet32x32. The better results obtained with snapshot ensembles confirm the complimentary benefits of warm restarts suggested by \citep{SnapshotICLR2017}. 

\subsection{Use of AdamW on other datasets and architectures}

Several other research groups have already successfully applied AdamW in citable works.
For example, \citet{2018arXiv180406559W} used AdamW to train a novel architecture for face detection on the standard WIDER FACE dataset~\citep{yang2016wider}, obtaining almost 10x faster predictions than the previous state of the art algorithms while achieving comparable performance. 
\citet{volker2018intracranial} employed AdamW with cosine annealing to train convolutional neural networks to classify and characterize error-related brain signals measured from intracranial electroencephalography (EEG) recordings.
%(24, one per patient). 
%While their paper does not provide a comparison to Adam, upon email inquiry, they kindly provided us with a direct comparison of the two on their best-performing problem-specific network architecture Deep4Net; the results demonstrate that AdamW yielded higher test set accuracy (73.68\%) than Adam (71.37\%) with the same hyperparameter setting. 
While their paper does not provide a comparison to Adam, they kindly provided us with a direct comparison of the two on their best-performing problem-specific network architecture Deep4Net and a variant of ResNet. AdamW  with the same hyperparameter setting as Adam yielded higher test set accuracy on Deep4Net (73.68\% versus 71.37\%) and statistically significantly higher test set accuracy on ResNet (72.04\% versus 61.34\%). 
\citet{radford2018improving} employed AdamW 
%our decoupled weight decay with Adam (AdamW) 
to train Transformer~\citep{vaswani2017attention} architectures to obtain new state-of-the-art results on a wide range of benchmarks for natural language understanding. \citet{zhang2018three} compared L$_2$ regularization vs.\ weight decay for SGD, Adam and the Kronecker-Factored Approximate Curvature (K-FAC) optimizer~\citep{martens2015optimizing} on the CIFAR datasets with ResNet and VGG architectures, reporting that decoupled weight decay consistently outperformed L$_2$ regularization in cases where they differ. 

\section{Conclusion and Future Work}

Following suggestions that adaptive gradient methods such as Adam might lead to worse generalization than SGD with momentum \citep{wilson2017marginal}, we identified and exposed % at least one possible explanation to this phenomenon: 
the inequivalence of L$_2$ regularization and weight decay for Adam. %We also identified why L$_2$ regularization is not efficient for Adam: Adam's normalization given by Eq. (\ref{eq:oradam}) renders regularization of weights with large parameter and/or gradient amplitudes insignificant (see Section \ref{sec:decoupling}).
We empirically showed that our version of Adam with decoupled weight decay %with the original formulation of weight decay 
yields substantially better generalization performance than the common implementation of Adam with L$_2$ regularization. We also proposed to use warm restarts for Adam to improve its anytime  performance. 

Our results obtained on image classification datasets must be verified on a wider range of tasks, especially ones where the use of regularization is expected to be important. It would be interesting to integrate our findings on weight decay into other methods which attempt to improve Adam, e.g, normalized direction-preserving Adam \citep{zhang2017normalized}. 
%FH: new sentence
While we focused our experimental analysis on Adam, we believe that similar results also hold for other adaptive gradient methods, such as AdaGrad~\citep{duchi2011adaptive} and AMSGrad ~\citep{reddi2018iclr}.

\section{Acknowledgments}

We thank Patryk Chrabaszcz for help with running experiments with ImageNet32x32; Matthias Feurer and Robin Schirrmeister for providing valuable feedback on this paper in several iterations; and Martin V\"{o}lker, Robin Schirrmeister, and Tonio Ball for providing us with a comparison of AdamW and Adam on their EEG data. 
We also thank the following members of the deep learning community for implementing decoupled weight decay in various deep learning libraries: 
\newcommand{\denselist}{\itemsep 0pt\partopsep -20pt}
\begin{itemize}
\denselist
\item Jingwei Zhang, Lei Tai, Robin Schirrmeister, and Kashif Rasul for their implementations in PyTorch (see \url{https://github.com/pytorch/pytorch/pull/4429})
\item Phil Jund for his implementation in TensorFlow described at\\ \url{https://www.tensorflow.org/api_docs/python/tf/contrib/opt/DecoupledWeightDecayExtension} 
\item Sylvain Gugger, Anand Saha, Jeremy Howard and other members of fast.ai for their implementation available at \url{https://github.com/sgugger/Adam-experiments}
\item Guillaume Lambard for his implementation in Keras available at \url{ https://github.com/GLambard/AdamW_Keras  }
\item Yagami Lin for his implementation in Caffe available at \url{ https://github.com/Yagami123/Caffe-AdamW-AdamWR  }
\end{itemize}

This work was supported by the European Research Council (ERC) under the European Union's Horizon 2020 research and innovation programme under grant no.\ 716721, by the German Research Foundation (DFG) under the BrainLinksBrainTools Cluster of Excellence (grant number EXC 1086) and through grant no.\ INST 37/935-1 FUGG, and by the German state of Baden-W\"{u}rttemberg through bwHPC. 

\bibliography{iclr2017_conference}
\bibliographystyle{iclr2018_conference}

\cleardoublepage
\setcounter{page}{1}

\appendix
{\begin{center}\Large{\textbf{Appendix}}\end{center}}
%\section{Proofs}\label{appendix:proofs}
\section{Formal Analysis of Weight Decay vs\ L$_2$ Regularization}
\label{sec:decay_vs_L_2}

%We now formalize the findings of section 2.%the previous section. 
%\begin{define}
%%[Adaptive gradient algorithm]
%An \emph{adaptive gradient algorithm} is an SGD variant with iterates on an objective function $f$ of
%\begin{eqnarray}
%	x_{t+1} \leftarrow x_t - \alpha \mathbf{M}_t \nabla f_{t}(\bm{\theta}_t), \label{eq:adagrad}
%\end{eqnarray}
%where $\mathbf{M}_t$ is a preconditioner matrix.
%%different from the identity matrix $I$.
%\end{define}

%\begin{prop}[Weight decay = L$_2$ reg for standard SGD]
%Standard SGD with base learning rate $\alpha$ executes the same steps on batch loss functions $f_t(\bm{\theta})$ with weight decay $\lambda$ (defined in Equation \ref{eq:wdecay}) as it executes without weight decay on $f_{t}^{\text{reg}}(\bm{\theta}) = f_t(\bm{\theta}) + \frac{\lambda'}{2} \norm{\bm{\theta}}_2^2$, with $\lambda' = \frac{\lambda}{\alpha}$.
%\end{prop}

\noindent{}\textbf{Proof of Proposition 1}\\
The proof for this well-known fact is straight-forward.
SGD without weight decay has the following iterates on $f_{t}^{\text{reg}}(\bm{\theta}) = f_t(\bm{\theta}) + \frac{\lambda'}{2} \norm{\bm{\theta}}_2^2$: 
\begin{equation}
\label{eq:sgdl2}
\bm{\theta}_{t+1} \leftarrow \bm{\theta}_t - \alpha \nabla f_{t}^{\text{reg}}(\bm{\theta}_t) = \bm{\theta}_t - \alpha \nabla f_t(\bm{\theta}_t) - \alpha  \lambda' \bm{\theta}_t.
\end{equation}
SGD with weight decay has the following iterates on $f_t(\bm{\theta})$: 
\begin{equation}
\label{eq:sgdwd}
\bm{\theta}_{t+1} \leftarrow (1 - \lambda) \bm{\theta}_t - \alpha \nabla f_t(\bm{\theta}_t).
\end{equation}
These iterates are identical since $\lambda' = \frac{\lambda}{\alpha}$. \qed

%\begin{prop}[Weight decay $\neq$ L$_2$ reg for adaptive gradients] Let $O$ denote an optimizer that has iterates $\bm{\theta}_{t+1} \leftarrow \bm{\theta}_t - \alpha \mathbf{M}_t \nabla f_t(\bm{\theta}_t)$ when run on batch loss function $f_t(\bm{\theta})$ \emph{without} weight decay, and 
%$\bm{\theta}_{t+1} \leftarrow (1 - \lambda) \bm{\theta}_t - \alpha \mathbf{M}_t \nabla f_t(\bm{\theta}_t)$ when run on $f_t(\bm{\theta})$ \emph{with} weight decay, respectively, with $\mathbf{M}_t \neq k \mathbf{I}$ (where $k\in\mathbb{R}$). 
%
%Then, for $O$ there exists no L$_2$ coefficient $\lambda'$ such that running $O$ on batch loss $f^{\text{reg}}_t(\bm{\theta}) = f_t(\bm{\theta}) + \frac{\lambda'}{2} \norm{\bm{\theta}}_2^2$ without weight decay is equivalent to running $O$ on $f_t(\bm{\theta})$ with decay $\lambda\in\mathbb{R}^+$. 
%%For adaptive gradient algorithms with $\mathbf{M}_t \neq k \mathbf{I}$ (where $k\in\mathrm{R}$), there exists no L$_2$ regularizer $\lambda' \norm{\bm{\theta}}_2^2$ that is equivalent to the weight decay regularization defined in Equation \ref{eq:wdecay}.
%\end{prop}

\noindent{}\textbf{Proof of Proposition 2}\\
Similarly to the proof of Proposition 1, the iterates of $O$ without weight decay on $f^{\text{reg}}_t(\bm{\theta}) = f_t(\bm{\theta}) + \frac{1}{2}  \lambda' \norm{\bm{\theta}}_2^2$ and $O$ with weight decay $\lambda$ on $f_t$ are, respectively:
\begin{eqnarray}
\label{eq:adgrl2}
\bm{\theta}_{t+1} &\leftarrow& \bm{\theta}_t - \alpha  \lambda' \mathbf{M}_t \bm{\theta}_t - \alpha \mathbf{M}_t \nabla f_t(\bm{\theta}_t).\\
\label{eq:adgr_wd}
\bm{\theta}_{t+1} &\leftarrow& (1 - \lambda) \bm{\theta}_t - \alpha \mathbf{M}_t \nabla f_t(\bm{\theta}_t).
\end{eqnarray}
The equality of these iterates for all $\bm{\theta}_t$ would imply
$\lambda\bm{\theta}_t = \alpha  \lambda' \mathbf{M}_t\bm{\theta}_t$.
This can only hold for all $\bm{\theta}_t$ if $\mathbf{M}_t = k \mathbf{I}$, with $k\in\mathbb{R}$, which is not the case for $O$. Therefore, no L$_2$ regularizer $\lambda' \norm{\bm{\theta}}_2^2$ exists that makes the iterates equivalent.
\qed

\noindent{}\textbf{Proof of Proposition 3}\\
$O$ without weight decay has the following iterates on $f_{t}^{\text{sreg}}(\bm{\theta}) = f_t(\bm{\theta}) + \frac{\lambda'}{2} \norm{\bm{\theta} \odot{} \sqrt{\vc{s}}}_2^2$: 
\begin{eqnarray}
\label{eq:sgdl2}
\bm{\theta}_{t+1} & \leftarrow &\bm{\theta}_t - \alpha \nabla f_{t}^{\text{sreg}}(\bm{\theta}_t)/\vc{s}\\
&=& \bm{\theta}_t - \alpha \nabla f_t(\bm{\theta}_t)/\vc{s} - \alpha  \lambda' \bm{\theta}_t \odot \vc{s} / \vc{s}\\
&=& \bm{\theta}_t - \alpha \nabla f_t(\bm{\theta}_t)/\vc{s} - \alpha  \lambda' \bm{\theta}_t,
\end{eqnarray}
where the division by $\vc{s}$ is element-wise.
$O$ with weight decay has the following iterates on $f_t(\bm{\theta})$: 
\begin{eqnarray}
\bm{\theta}_{t+1} &\leftarrow& (1 - \lambda) \bm{\theta}_t - \alpha \nabla f(\bm{\theta}_t) / \vc{s}\\
&=& \bm{\theta}_t - \alpha \nabla f(\bm{\theta}_t) / \vc{s} -\lambda \bm{\theta}_t ,
\end{eqnarray}
These iterates are identical since $\lambda' = \frac{\lambda}{\alpha}$. \qed

\section{Additional Practical Improvements of Adam}\label{sec:practical_improvements}
Having discussed decoupled weight decay for improving Adam's generalization, in this section we introduce two additional components to improve Adam's performance in practice.

\subsection{Normalized Weight Decay}
\label{sec:normw}
Our preliminary experiments showed that different weight decay factors are optimal for different computational budgets (defined in terms of the number of batch passes).
Relatedly, \citet{li2017visualizing} demonstrated that a smaller batch size (for the same total number of epochs) leads to the shrinking effect of weight decay being more pronounced.
Here, we propose to reduce this dependence by normalizing the values of weight decay. Specifically, we replace the hyperparameter $\lambda$ by a new (more robust) normalized weight decay hyperparameter $\lambda_{norm}$, and use this to set $\lambda$ 
as $\lambda = \lambda_{norm} \sqrt{ \frac{b}{B T} }$,	 
%as follows:
%\begin{eqnarray}
%	\label{eq:wnorm}
%    \lambda= \lambda_{norm} \sqrt{ \frac{b}{B T} },	
%\end{eqnarray}
where $b$ is the batch size, $B$ is the total number of training points and $T$ is the total number of epochs.\footnote{In the context of our AdamWR variant discussed in Section \ref{sec:adamwr}, $T$ is the total number of epochs in the current restart.}
%within the $i$-th run/restart of the algorithm. 
Thus, $\lambda_{norm}$ can be interpreted as the weight decay used if only one batch pass is allowed.
\franknips{We emphasize that our choice of normalization is merely one possibility informed by few experiments; a more lasting conclusion we draw is that using \emph{some} normalization can substantially improve results.}
%While this normalization works well in practice
%We would like to mention that the particular normalization we chose is just one of many possibilities; better ones may exist.
%; the important thing is to consider \emph{some} normalization.
%
%as we only developed it based on one dataset (CIFAR-10) and verified it based on one additional dataset (ImageNet32x32). It is 
%\textbf{can be interpreted as the weight decay to be used if only one batch pass is allowed}. 
%Our proposal is motivated by the observation that the original way to apply the weight decay regularization is sensitive to the choice of the batch size and the total number of epochs because they will effectively define the total number of iterations and thus the `total'  decay. Instead, we hypothesize that one can define the `total' amount of the weight decay and then depending on the choice of the batch size and the total number of epochs, the actual weight decay factor for each batch can be computed. Thus, one can explicitly decouple this hyperparameter setting from the choice of $b$ and $T_i$, however, some problem-dependent implicit coupling is likely to remain. This decoupling might be useful to deal with the cases when $b$ and / or  $T_i$ are not fixed, e.g., when $T_i$ changes at each restart as in SGDR, AdamR and AdamWR. 

\subsection{Adam with Cosine Annealing and Warm Restarts}
% and Normalized Weight Decay
\label{sec:adamwr}

We now apply cosine annealing and warm restarts to Adam, following 
% de-anonymized, anonymize it again as follows
%the recent work of \citet{loshchilov2016sgdr}. There, the authors 
our recent work \citep{loshchilov2016sgdr}. There, we
proposed Stochastic Gradient Descent with Warm Restarts (SGDR) to improve the anytime performance of SGD by quickly cooling down the learning rate according to a cosine schedule and periodically increasing it. 
SGDR has been successfully adopted to lead to new state-of-the-art results for popular image classification benchmarks \citep{SnapshotICLR2017,gastaldi2017shake,zoph-arxiv17b}, and we therefore already tried extending it to Adam shortly after proposing it. 
However, while our initial version of Adam with warm restarts had better anytime performance than Adam, it was not competitive with SGD with warm restarts, precisely because L$_2$ regularization was not working as well as in SGD.
%(inherited by the version with warm restarts). 
Now, having fixed this issue by means of the original weight decay regularization (Section \ref{sec:decoupling}) and also having introduced normalized weight decay (Section \ref{sec:normw}), 
our original work on cosine annealing and warm restarts
%the original work on cosine annealing and warm restarts by \citet{loshchilov2016sgdr} 
directly carries over to Adam.
%, and we use it to construct AdamWR to fully benefit from warm restarts. 

%Naturally, we attempted to employ the same approach for Adam shortly afterwards, but while our initial version of Adam with warm restarts improved anytime performance of Adam it was not competitive to SGD with warm restarts precisely because of Adam's issues (see section 2) that the version with warm restarts inherited.  Now, having fixed weight decay regularization and also having introduced normalized weight decay, our original work on warm restarts directly carries over, and we use it to construct AdamWR to fully benefit from warm restarts. 

%In our recent work \citep{loshchilov2016sgdr}, we proposed Stochastic Gradient Descent with Warm Restarts (SGDR) to improve anytime performance of SGD by periodically increasing the learning rate. Naturally, we attempted to employ the same approach for Adam. While our initial version of Adam with warm restarts improved anytime performance of Adam, the approach was not competitive to SGD with warm restarts precisely because of the Adam's issues (see section 2) that the version with warm restarts inherited.  Now, having fixed weight decay regularization and also having introduced normalized weight decay, our original work on warm restarts directly carries over, and we use it to construct AdamWR to fully benefit from warm restarts.  

%\note{FH: a lot of this could probably be moved to the appendix; by now cosine annealing is not so new anymore.}
In the interest of keeping the presentation self-contained, we briefly describe how SGDR schedules the change of the effective learning rate in order to accelerate the training of DNNs. Here, we decouple the initial learning rate $\alpha$ and its multiplier $\eta_t$ used to obtain the actual learning rate at iteration $t$ (see, e.g., line \ref{sgd-mom1} in Algorithm 1). 
In SGDR, we simulate a new warm-started run/restart of SGD once $T_i$ epochs are performed, where $i$ is the index of the run. Importantly, the restarts are not performed from scratch but emulated by increasing $\eta_t$ while the old value of $\bm{\theta}_{t}$ is used as an initial solution. The amount by which $\eta_t$ is increased controls to which extent the previously acquired information (e.g., momentum) is used. Within the $i$-th run, the value of $\eta_t$ decays according to a cosine annealing~\citep{loshchilov2016sgdr} 
learning rate for each batch as follows:
\begin{eqnarray}
	\label{eq:t}
	\eta_t = \eta^{(i)}_{min} + 0.5 (\eta^{(i)}_{max} - \eta^{(i)}_{min}) (1 + \cos(\pi T_{cur} / {T_i})),	
\end{eqnarray}
where $\eta^{(i)}_{min}$ and $\eta^{(i)}_{max}$ are ranges for the multiplier and $T_{cur}$ accounts for how many epochs have been performed since the last restart.  $T_{cur}$ is updated at each batch iteration $t$ and is thus not constrained to integer values.
%FH: was "can take discretized values such as 0.1, 0.2, etc." 
%FH: removed the repetitive self-reference in this non-blind paper. Was:
%As discussed by  \cite{loshchilov2016sgdr}, one can adjust (e.g., decrease) $\eta^{(i)}_{min}$ and $\eta^{(i)}_{max}$ at every $i$-th restart (see also \cite{smith2016}). While this could potentially improve the performance, following \cite{loshchilov2016sgdr}, we do not consider that option in our experiments because it would involve additional hyperparameters. 
Adjusting (e.g., decreasing) $\eta^{(i)}_{min}$ and $\eta^{(i)}_{max}$ at every $i$-th restart (see also \cite{smith2016}) could potentially improve performance, but we do not consider that option here because it would involve additional hyperparameters. 
For $\eta^{(i)}_{max}=1$ and $\eta^{(i)}_{min}=0$, one can simplify Eq. (\ref{eq:t}) to
\begin{eqnarray}
	\label{eq:t2}
	\eta_t = 0.5 + 0.5\cos(\pi T_{cur} / {T_i}).	
\end{eqnarray}
In order to achieve good anytime performance, one can start with an initially small $T_i$ (e.g., from 1\% to 10\% of the expected total budget) and multiply it by a factor of $T_{mult}$ (e.g., $T_{mult}=2$) at every restart. The $(i+1)$-th restart is triggered when $T_{cur} = T_i$ by setting $T_{cur}$ to 0. An example setting of the schedule multiplier is given in  \ref{sec:example}. 
%Note that the effective learning rate is controlled by $\eta_t \alpha$ where $\alpha$ is set to the initial learning rate and stays constant in our experimental setup.
%The reason why we employ $\alpha$ and not simply $\alpha$ is to account for possible practical extensions, e.g., to adapt $\alpha$ as a function of batch size in (scheduled) large-batch settings. 

Our proposed \textbf{AdamWR} algorithm represents AdamW (see Algorithm 2) with $\eta_t$ following Eq.\ (\ref{eq:t2}) and $\lambda$ computed at each iteration using normalized weight decay described in Section \ref{sec:normw}. % according to Eq.\ (\ref{eq:wnorm}). 
%FH: new sentence
We note that normalized weight decay allowed us to use a constant parameter setting across short and long runs performed within AdamWR and SGDWR (SGDW with warm restarts).
%FH: question: why didn't we need normalized weight decay for SGDR? Does it also make that better?
%Equivalently to AdamWR, we define SGDWR as SGDW with warm restarts.

\section{An Example Setting of the Schedule Multiplier}
\label{sec:example}

An example schedule of the schedule multiplier $\eta_t$ is given in SuppFigure \ref{fig1_lr} for $T_{i=0}=100$ and $T_{mult}=2$. After the initial 100 epochs the learning rate will reach 0 because $\eta_{t=100}=0$. Then, since $T_{cur} = T_{i=0}$, 
we restart by resetting $T_{cur}=0$, causing the multiplier $\eta_t$ to be reset to 1 due to Eq. (\ref{eq:t2}). This multiplier will then decrease again from 1 to 0, but now over the course of 200 epochs because $T_{i=1}=T_{i=0} T_{mult}=200$. Solutions obtained right before the restarts, when $\eta_t=0$ (e.g., at epoch indexes 100, 300, 700 and 1500 as shown in SuppFigure \ref{fig1_lr}) are recommended by the optimizer as the solutions, with more recent solutions prioritized.

\setcounter{figure}{0}
\makeatletter
\renewcommand{\fnum@figure}{SuppFigure ~\thefigure}
\makeatother

\begin{figure}[t!]%
	\center{\includegraphics[width=0.5\textwidth]{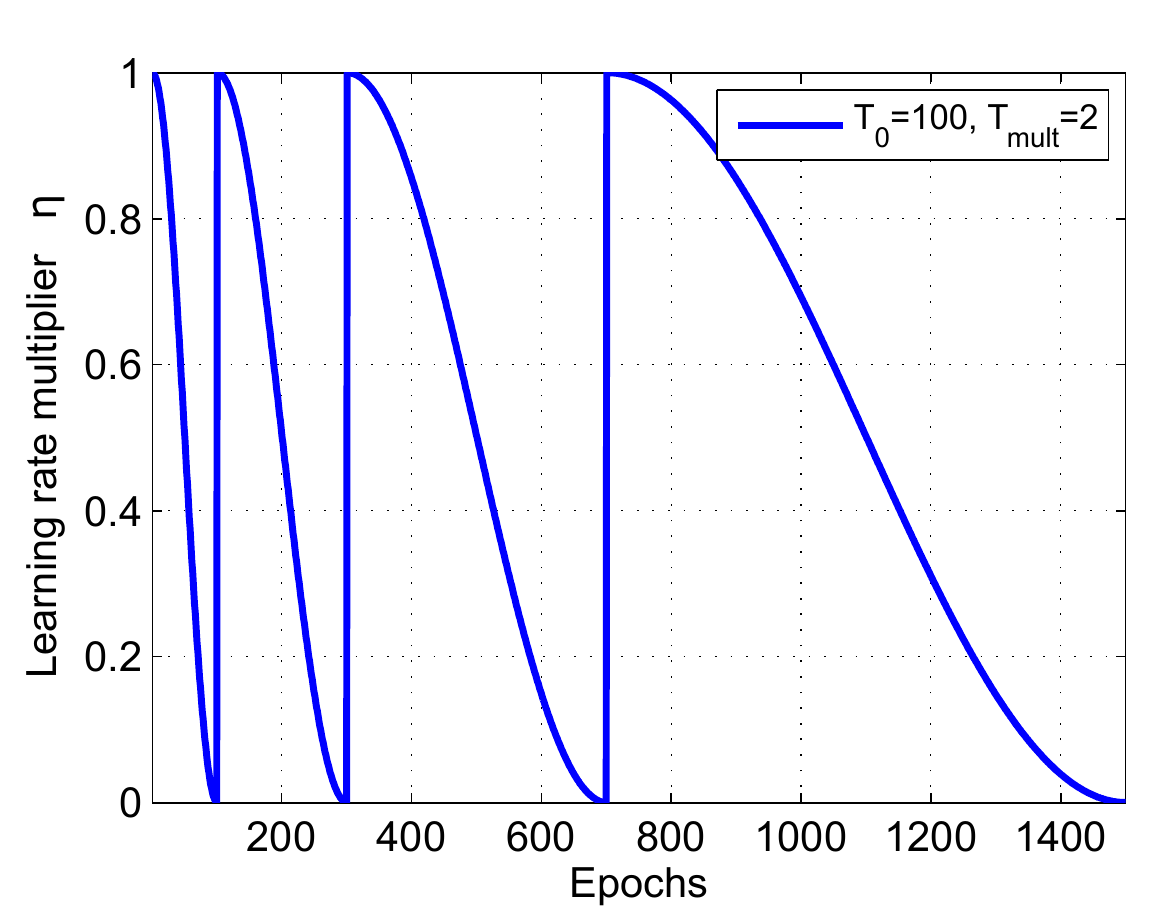}} 
\caption{\label{fig1_lr} An example schedule of the learning rate multiplier as a function of epoch index. The first run is scheduled to converge at epoch $T_{i=0}=100$, then the budget for the next run is doubled as $T_{i=1}=T_{i=0} T_{mult}=200$, etc.} 
\end{figure}

\section{Additional Results}
\label{app:additional_results}

%Figure \ref{fig:adam_with_without_cosine} demonstrates that Adam yields much better results with cosine annealing than with a fixed learning rate. For this reason, we schedule the learning rate with cosine annealing for all methods given in the paper.

\begin{figure}[tb]%
	\centering
    \includegraphics[width=0.47\textwidth]{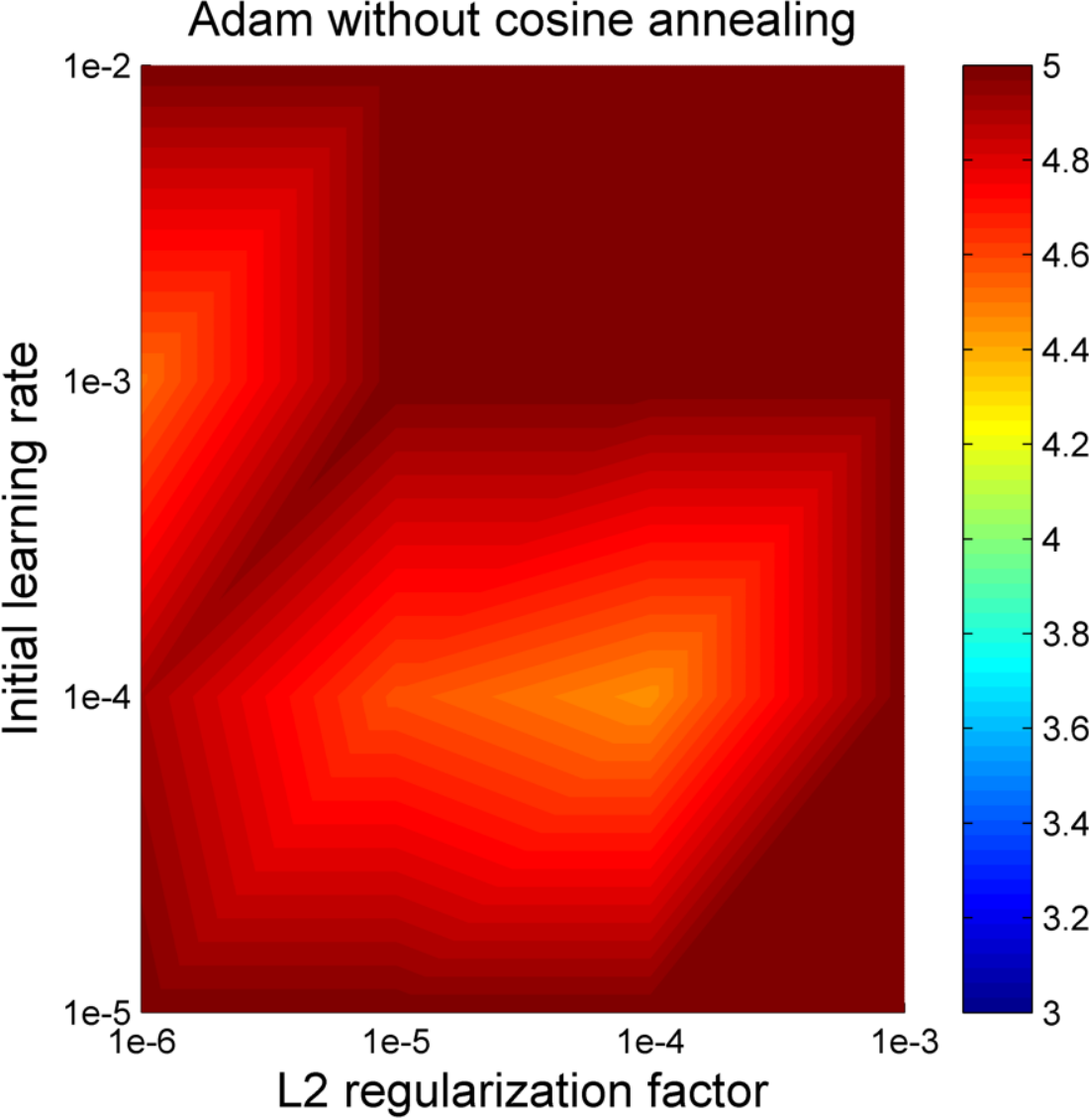} 
\caption{\label{baseline1800} Performance of ``standard Adam'': Adam with L$_2$ regularization and a fixed learning rate. We show the final test error of a 26 2x96d ResNet on CIFAR-10 after 1800 epochs of the original Adam for different settings of learning rate and weight decay used for L$_2$ regularization.}
\end{figure}

\begin{figure*}[p]%
\begin{center}
	\includegraphics[width=0.3\textwidth]{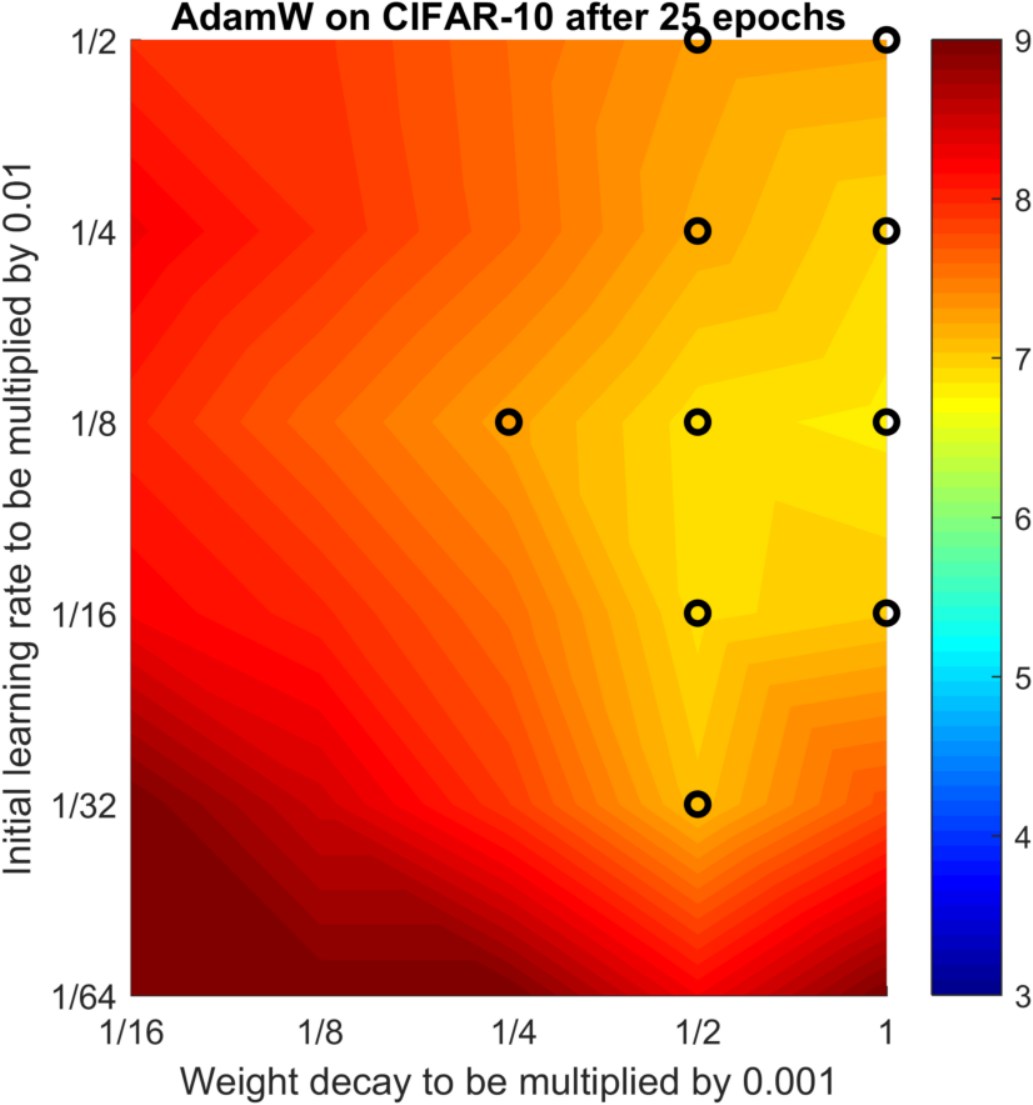} 
  \includegraphics[width=0.3\textwidth]{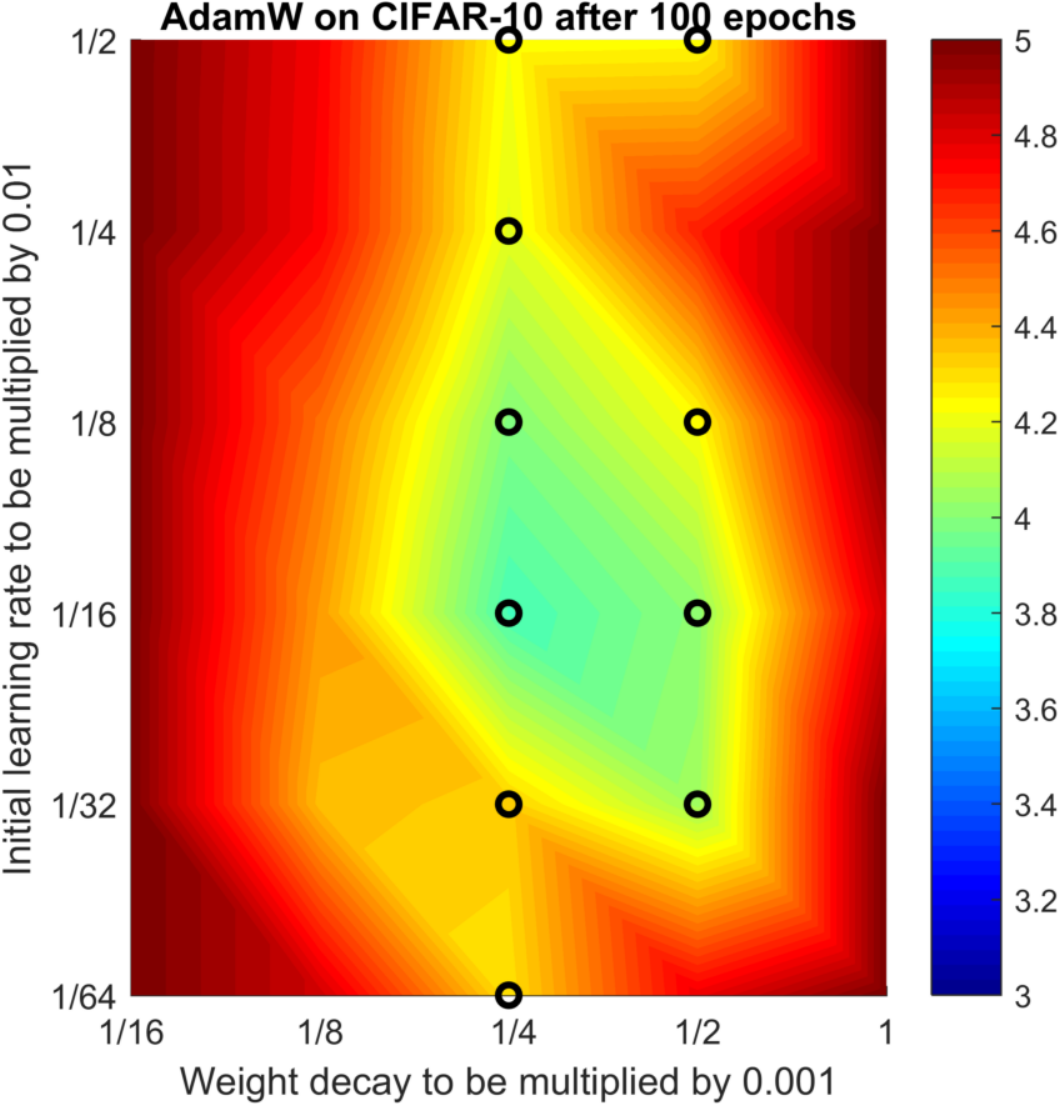} 
	\includegraphics[width=0.3\textwidth]{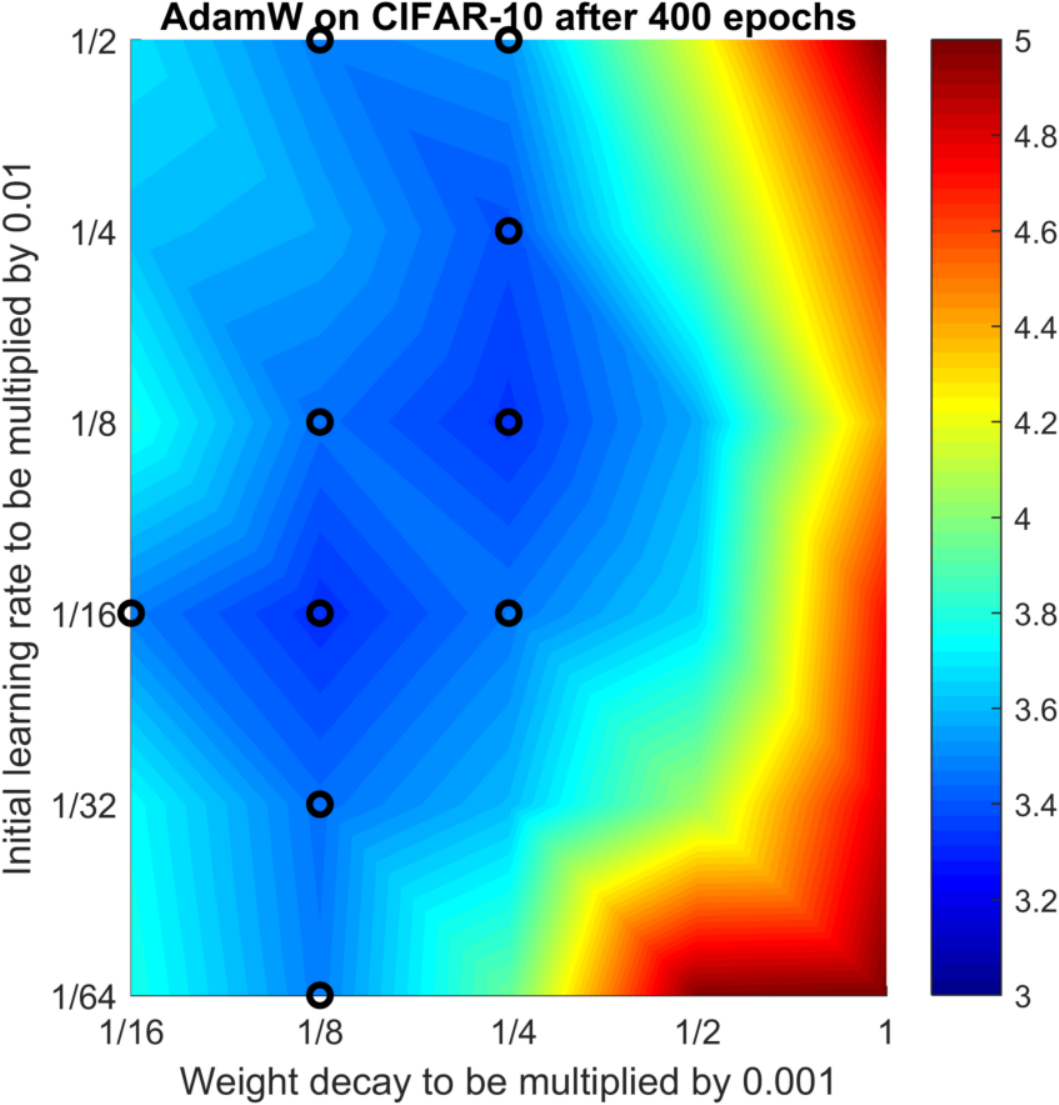}\\
	\includegraphics[width=0.3\textwidth]{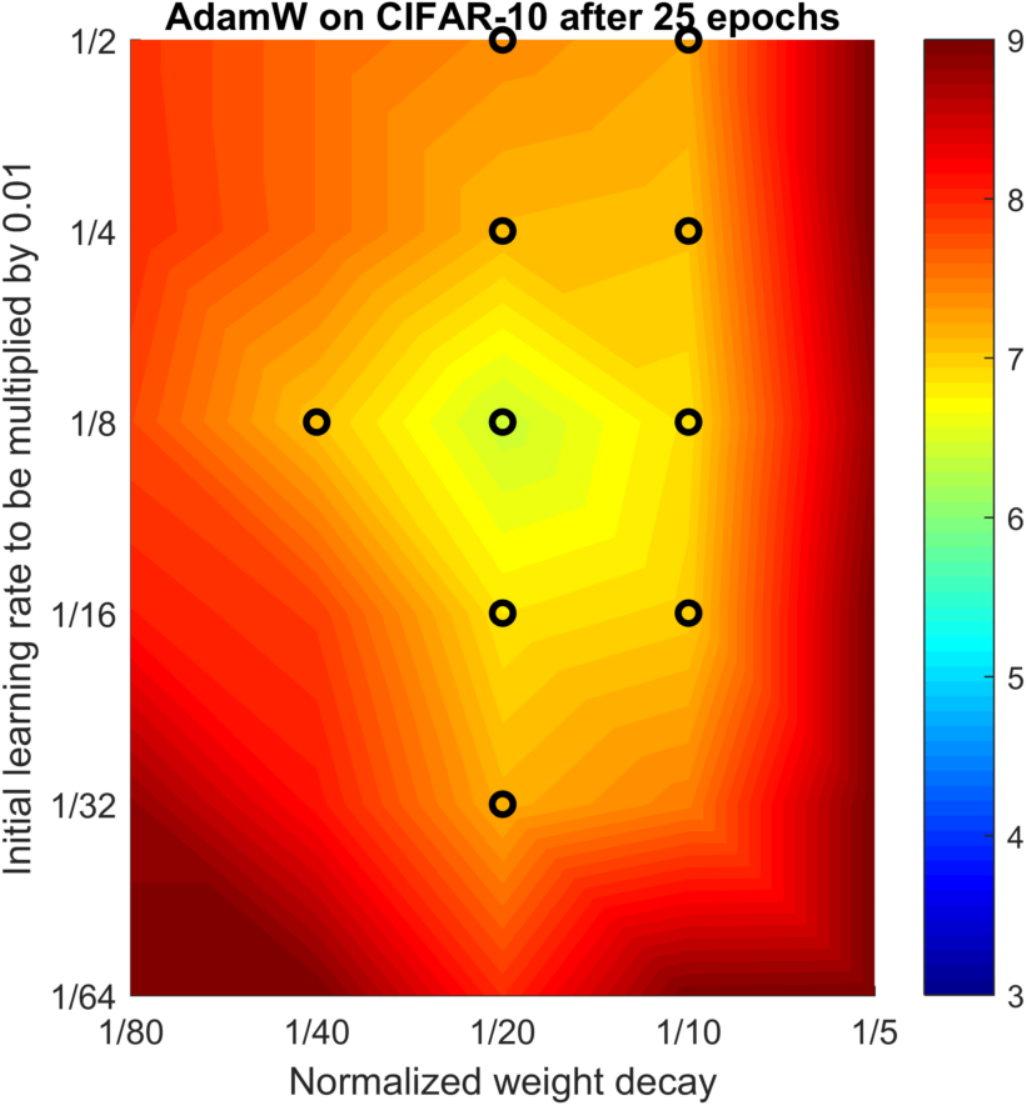} 
  \includegraphics[width=0.3\textwidth]{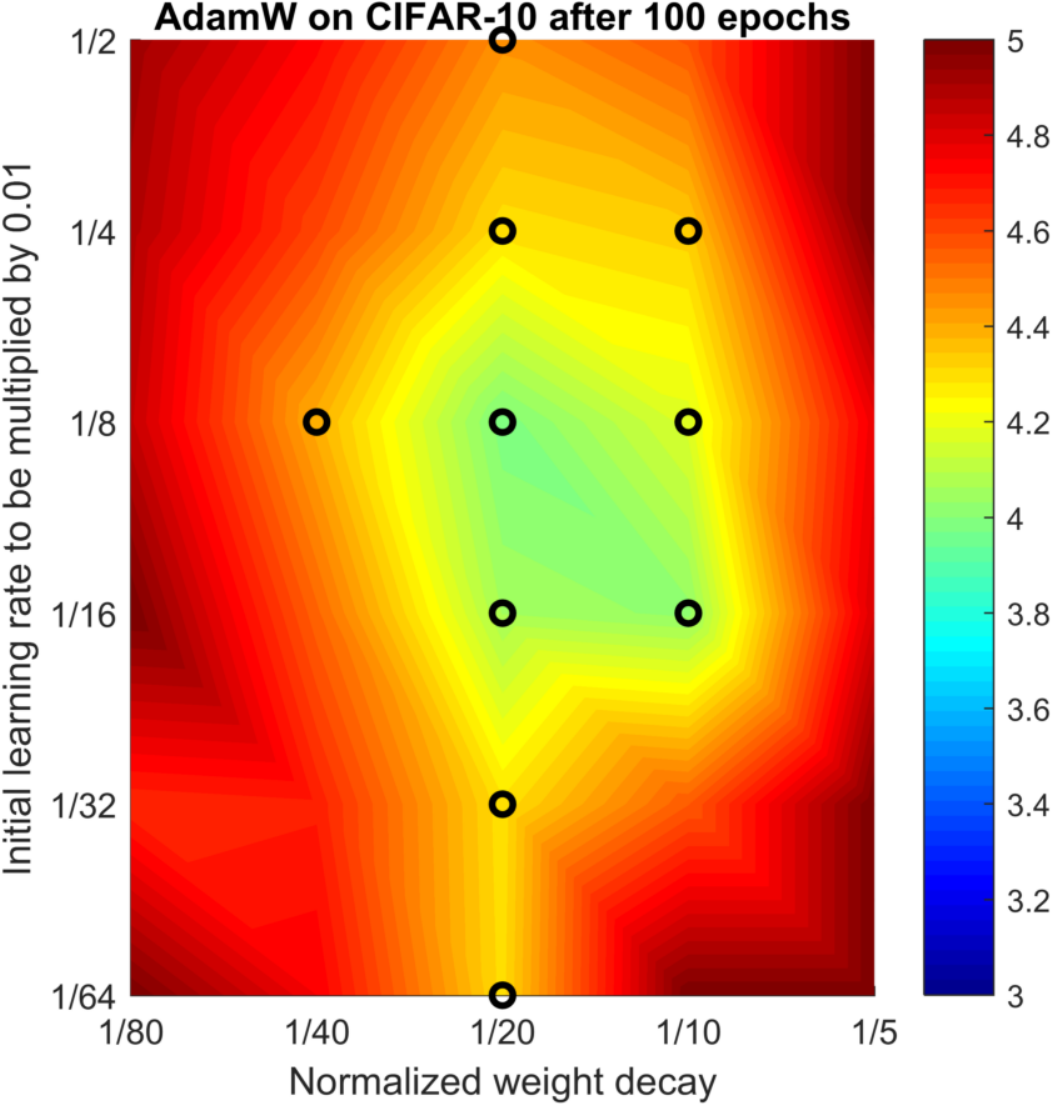}
  \includegraphics[width=0.3\textwidth]{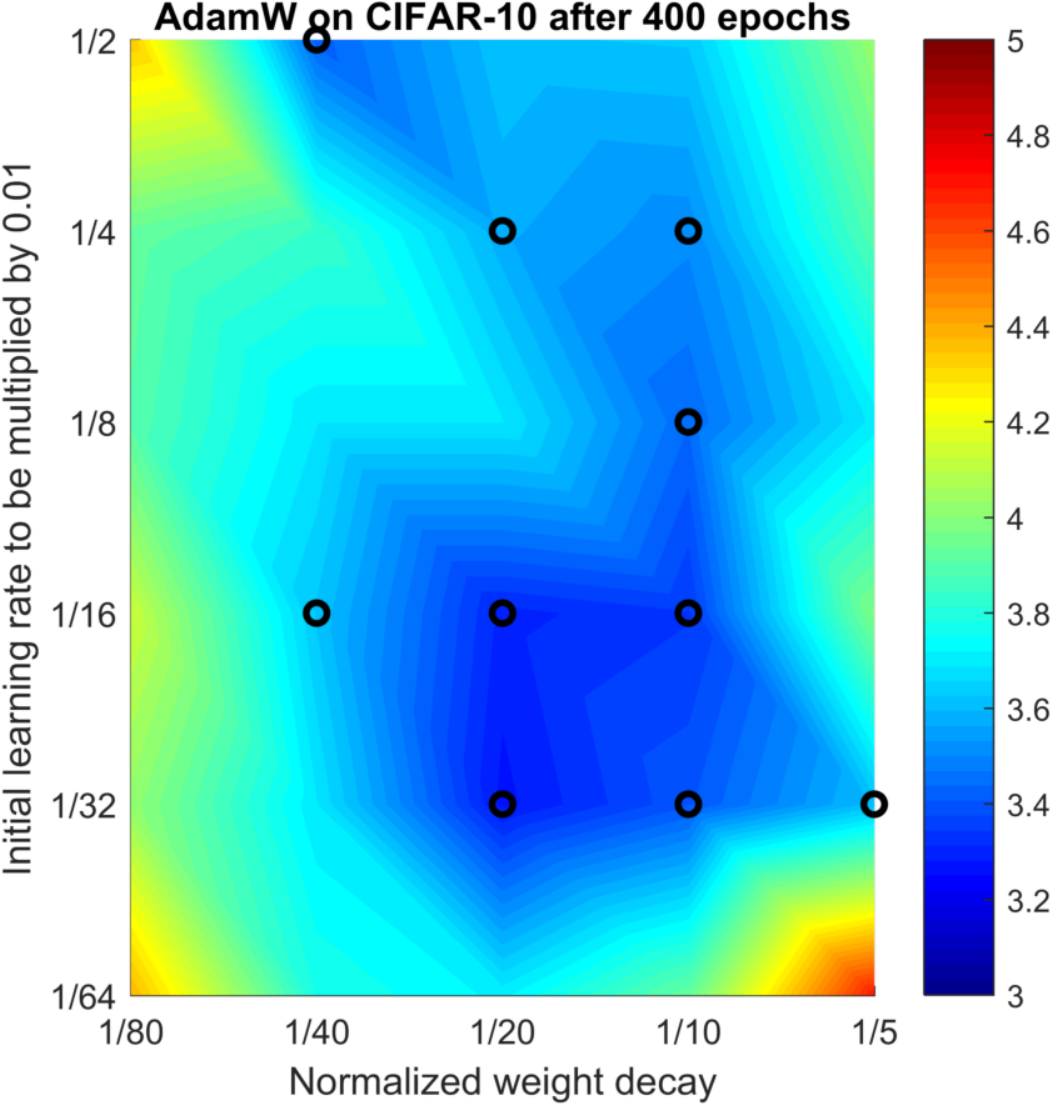}\\
 % \note{Do we also have plots for original Adam / original SGD on ImageNet 32x32? I think the clearest point about these plots is the contrast between the first and the second row.}
	\includegraphics[width=0.3\textwidth]{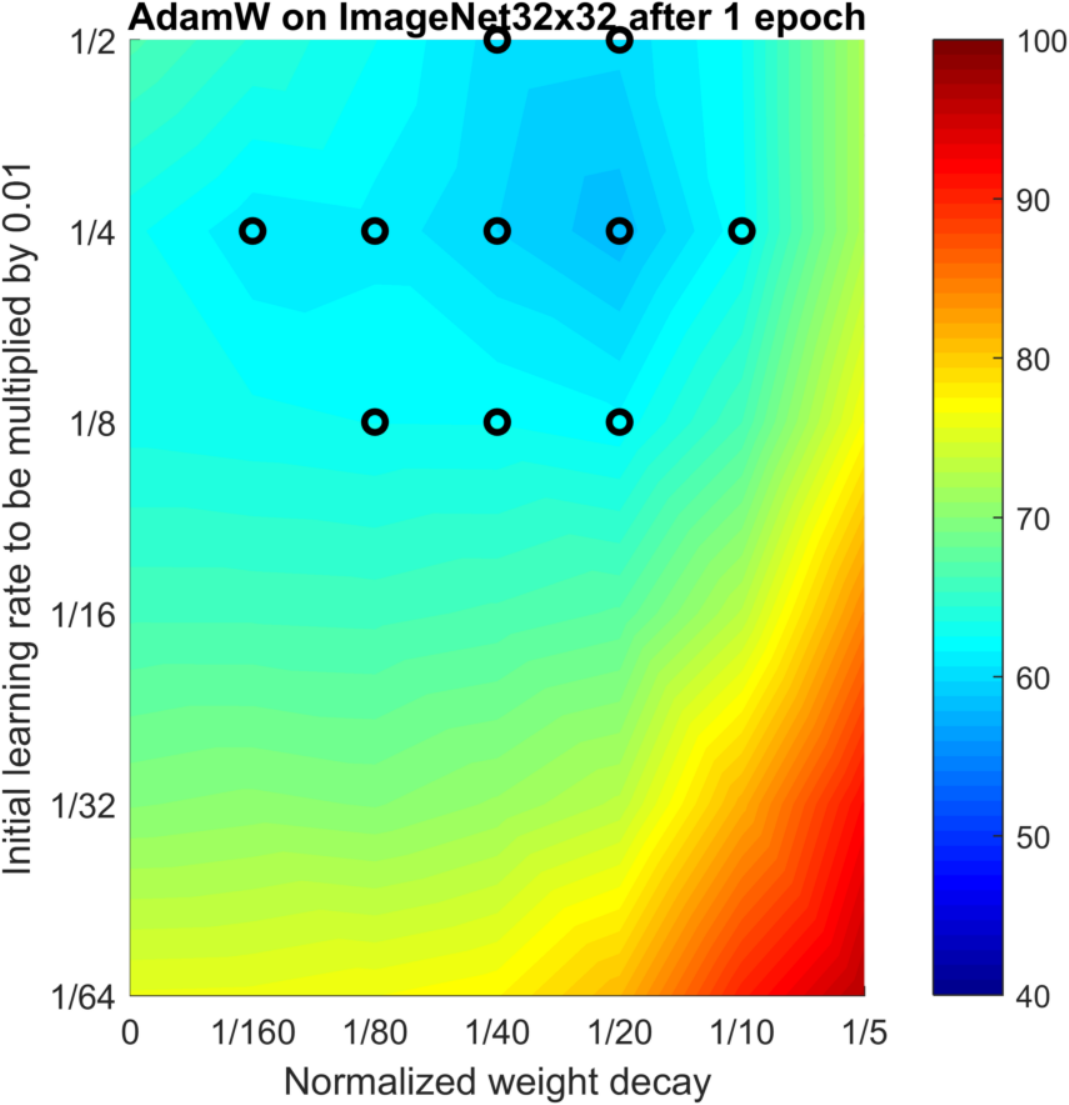}
  \includegraphics[width=0.3\textwidth]{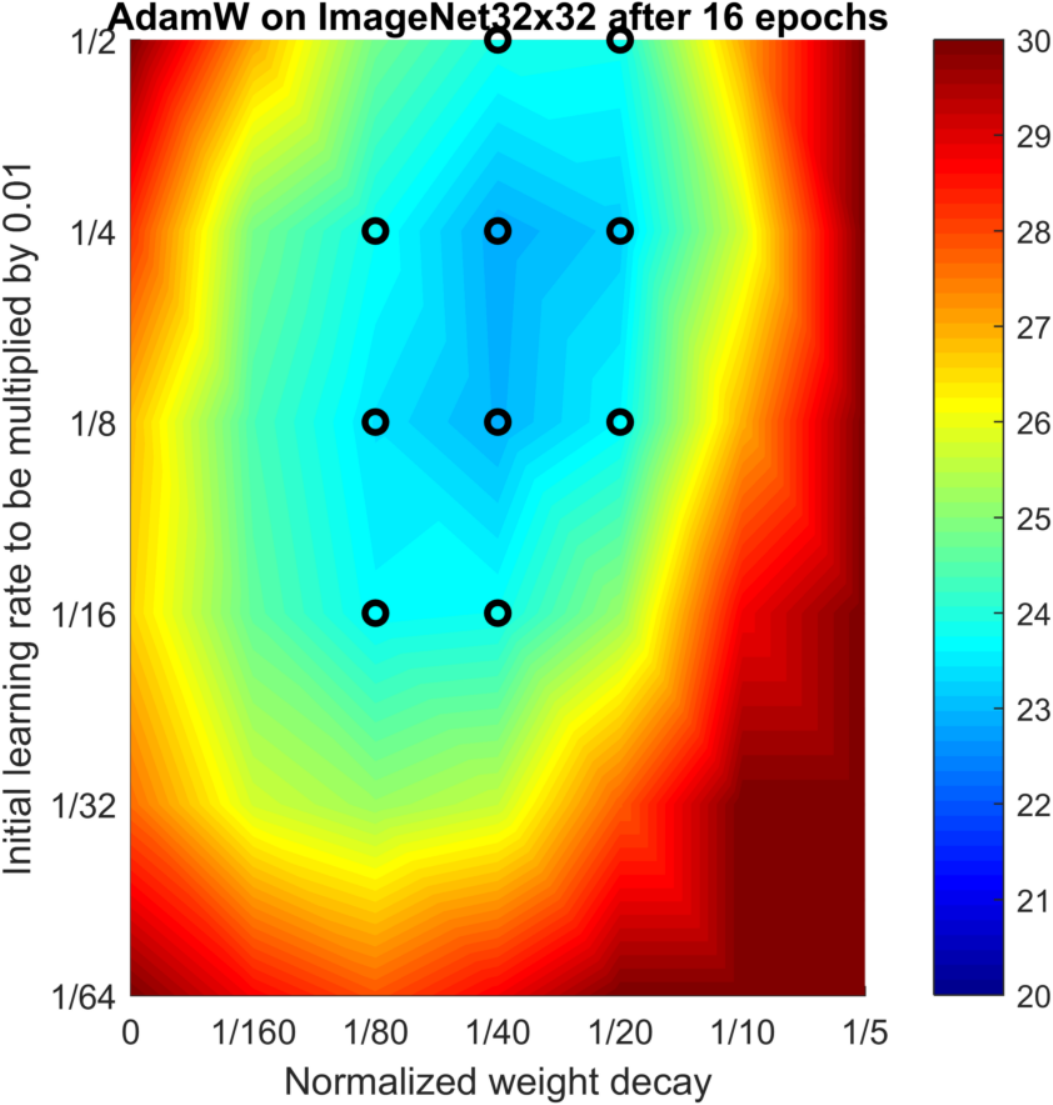}
  \includegraphics[width=0.3\textwidth]{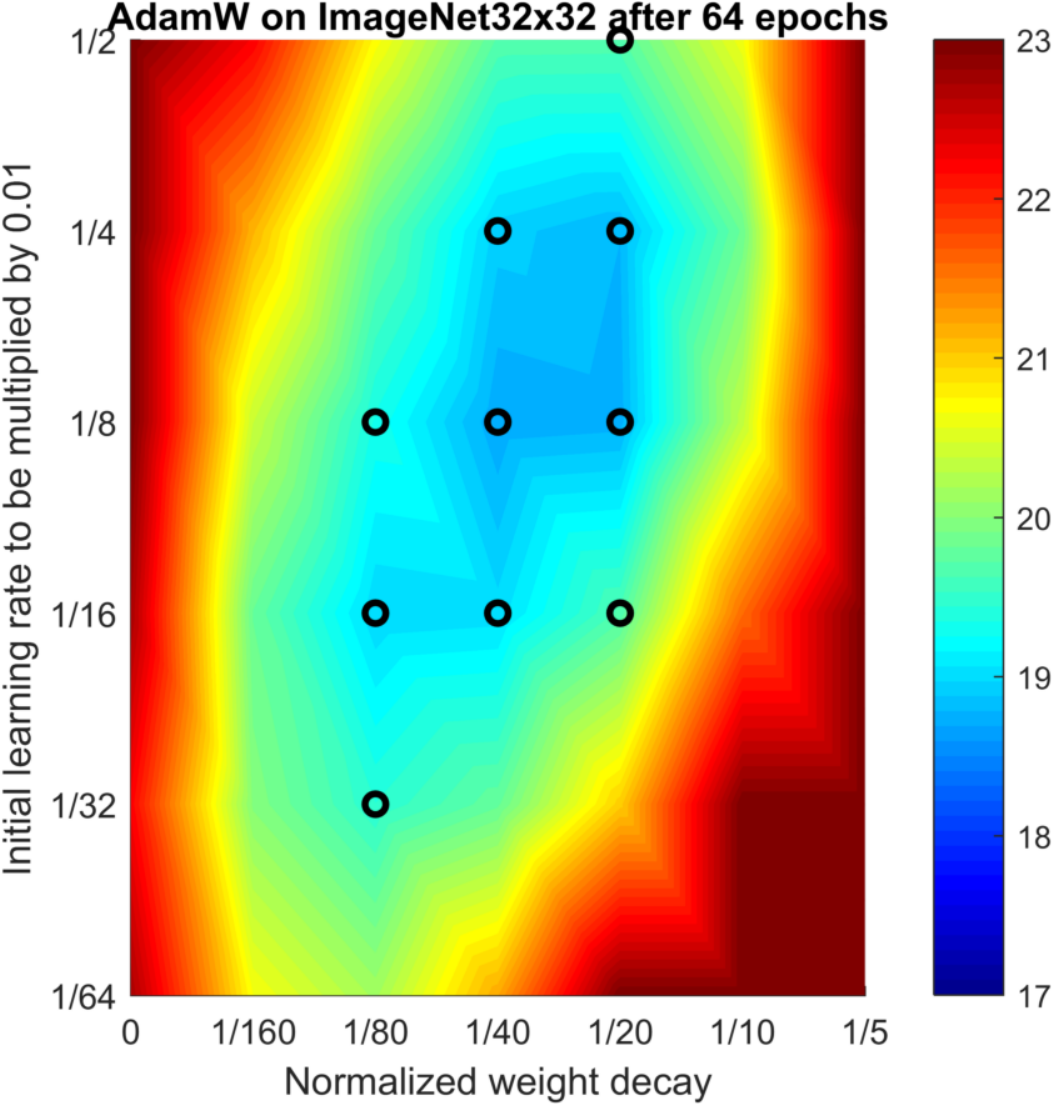}\\
  \includegraphics[width=0.3\textwidth]{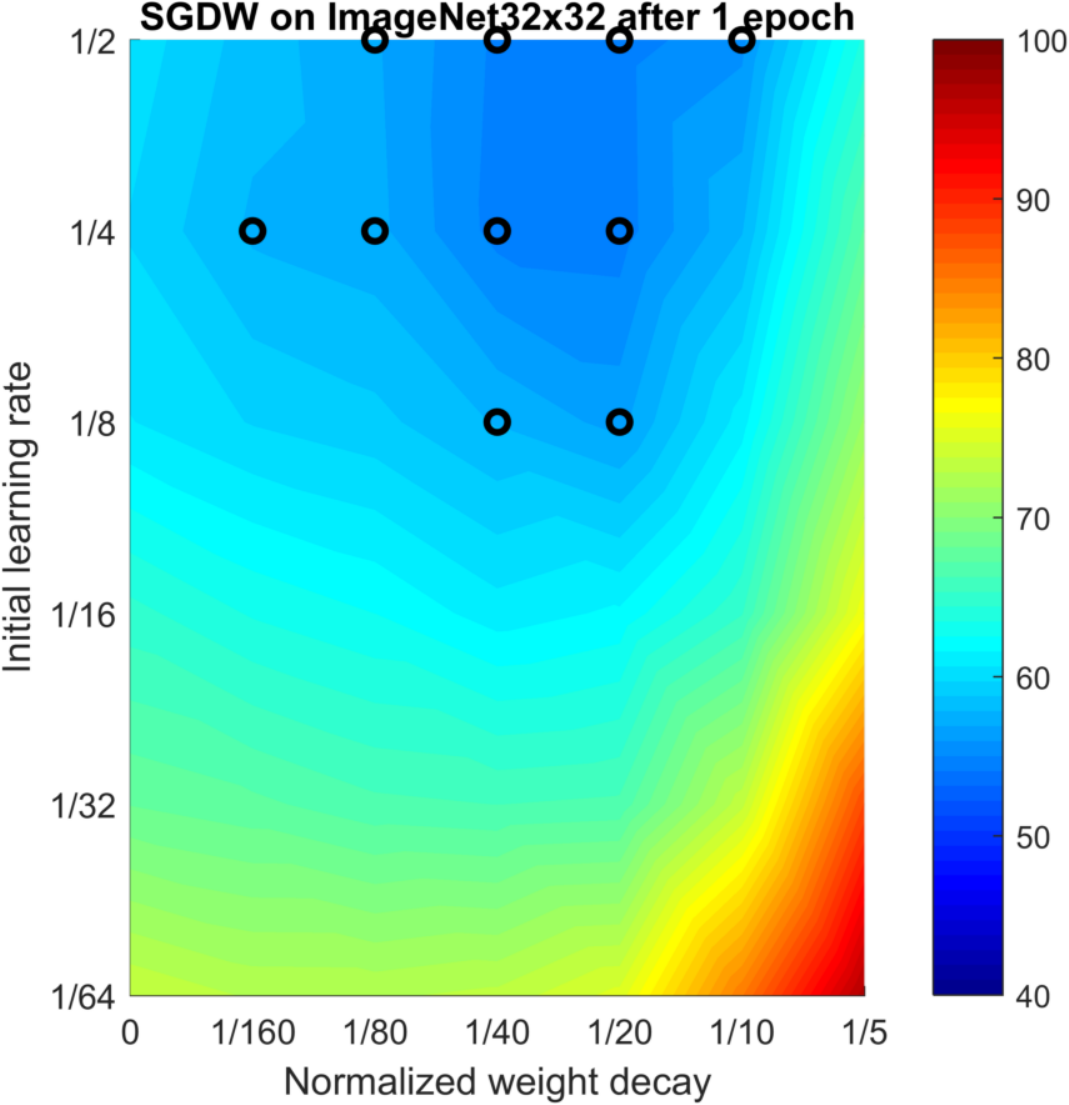}
  \includegraphics[width=0.3\textwidth]{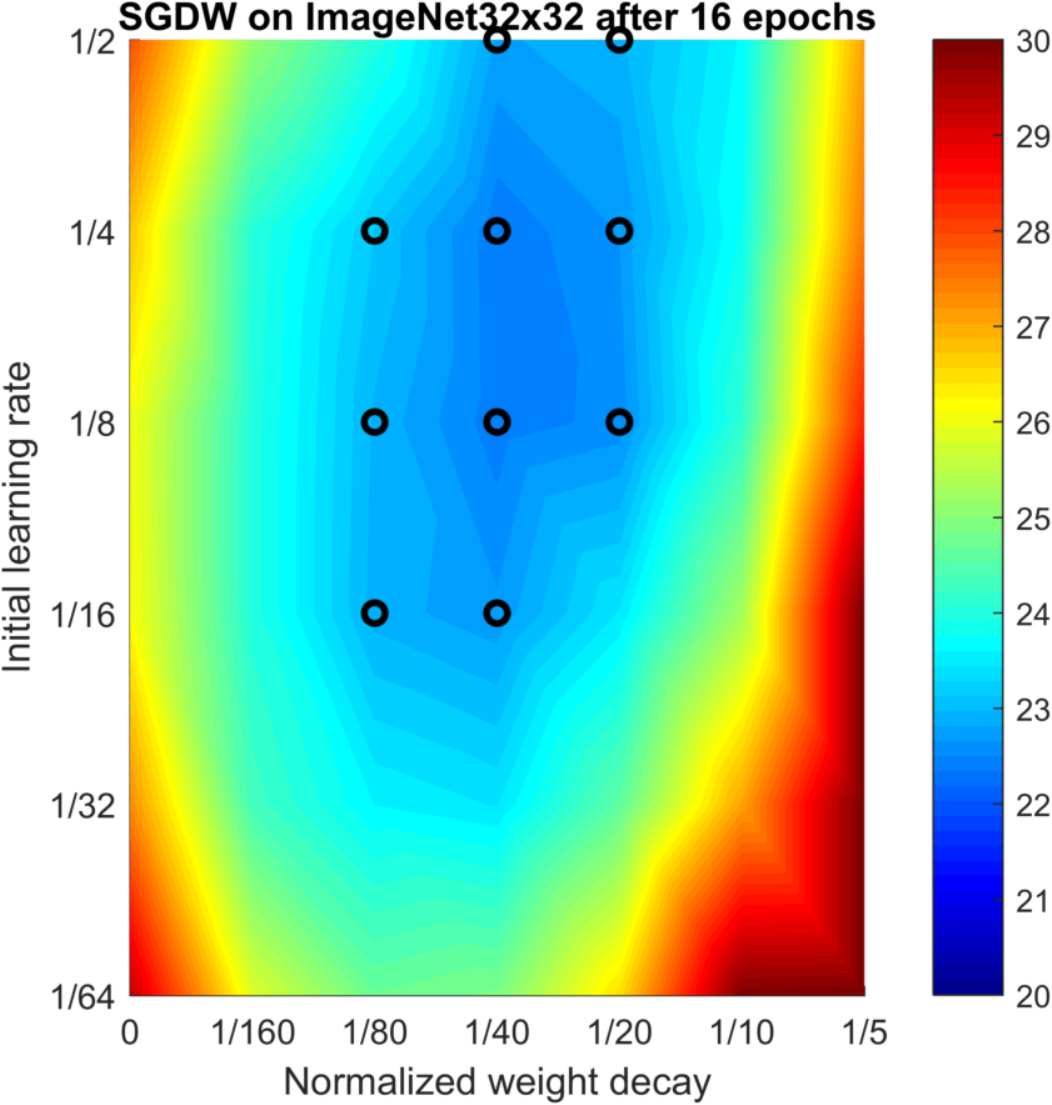}
  \includegraphics[width=0.3\textwidth]{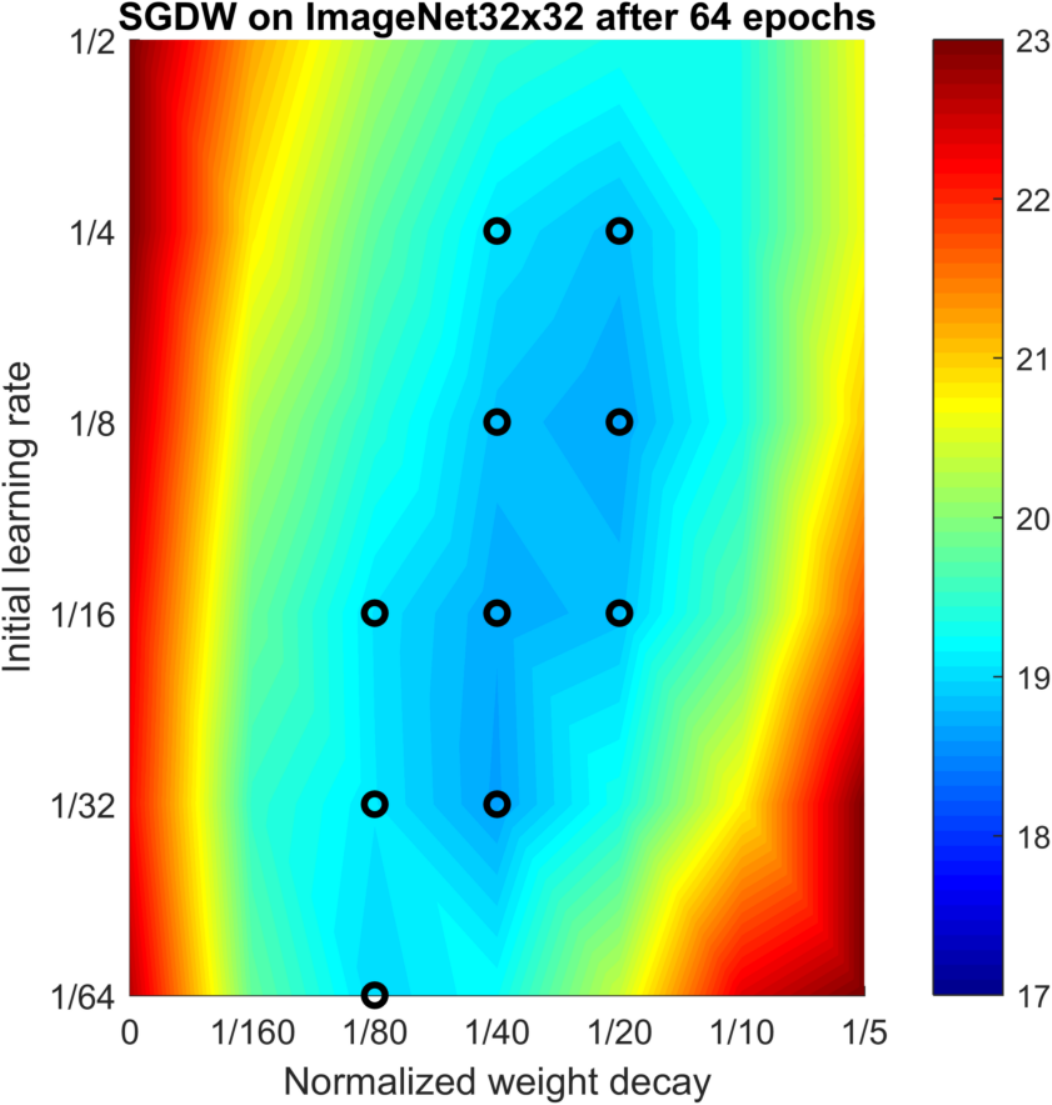}
\caption{\label{sfig_2} Effect of normalized weight decay. We show the final test Top-1 error on CIFAR-10 (first two rows for AdamW without and with normalized weight decay) and Top-5 error on ImageNet32x32 (last two rows for AdamW and SGDW, both with normalized weight decay) of a 26 2x64d ResNet after different numbers of epochs (see columns). While the optimal settings of the raw weight decay change significantly for different runtime budgets (see the first row), the values of the normalized weight decay remain very similar for different budgets (see the second row) and different datasets (here, CIFAR-10 and ImageNet32x32), and even across AdamW and SGDW.}
\end{center}
\end{figure*}

We investigated whether the use of much longer runs (1800 epochs) of ``standard Adam'' (Adam with L$_2$ regularization and a fixed learning rate) makes the use of cosine annealing unnecessary. SuppFigure \ref{baseline1800} shows the results of standard Adam for a 4 by 4 logarithmic grid of hyperparameter settings (the coarseness of the grid is due to the high computational expense of runs for 1800 epochs). Even after taking the low resolution of the grid into account, the results appear to be at best comparable to the ones obtained with AdamW with 18 times less epochs and a smaller network (see SuppFigure \ref{sfig_2}, top row, middle). These results are not very surprising given Figure 1 in the main paper (which demonstrates both the improvements possible by using some learning rate schedule, such as cosine annealing, and the effectiveness of decoupled weight decay).

Our experimental results with Adam and SGD suggest that the total runtime in terms of the number of epochs affect the basin of optimal hyperparameters (see SuppFigure \ref{sfig_2}). 
More specifically, the greater the total number of epochs the smaller the values of the weight decay should be. 
SuppFigure 4
%\ref{sfig_2} 
shows that our remedy for this problem, the normalized weight decay defined in Eq. (\ref{eq:t2}), simplifies hyperparameter selection because the optimal values observed for short runs are similar to the ones for much longer runs. 
We used our initial experiments on CIFAR-10 to suggest the square root normalization we proposed in Eq. (\ref{eq:t2}) and double-checked that this is not a coincidence on the ImageNet32x32 dataset \citep{chrabaszcz2017downsampled}, a downsampled version of the original ImageNet dataset with 1.2 million  32$\times$32 pixels images, where an epoch is 24 times longer than on CIFAR-10. This experiment also supported the square root scaling:  the best values of the normalized weight decay observed on CIFAR-10 represented nearly optimal values for ImageNet32x32 (see SuppFigure \ref{sfig_2}).
%\ref{sfig_2}). 
In contrast, had we used the same raw weight decay values $\lambda$ for ImageNet32x32 as for CIFAR-10 and for the same number of epochs, \emph{without the proposed normalization, $\lambda$ would have been roughly 5 times too large for ImageNet32x32, leading to much worse performance}. 
The optimal normalized weight decay values were also very similar (e.g., $\lambda_{norm} = 0.025$ and $\lambda_{norm} = 0.05$) across SGDW and AdamW.
These results clearly show that normalizing weight decay can substantially improve performance; while square root scaling performed very well in our experiments we emphasize that these experiments were not very comprehensive and that even better scaling rules are likely to exist. 

%SuppFigure \ref{sfig_2} demonstrates the effects of normalized weight decay: the optimal setting of weight decay differs substantially for different runtime budgets (see the first row), the values for normalized weight decay remain very similar (see the second row). They also remain similar across datasets and even across AdamW and SGDW (see the third and fourth rows).

\begin{figure*}[t]%
	\includegraphics[width=0.47\textwidth]{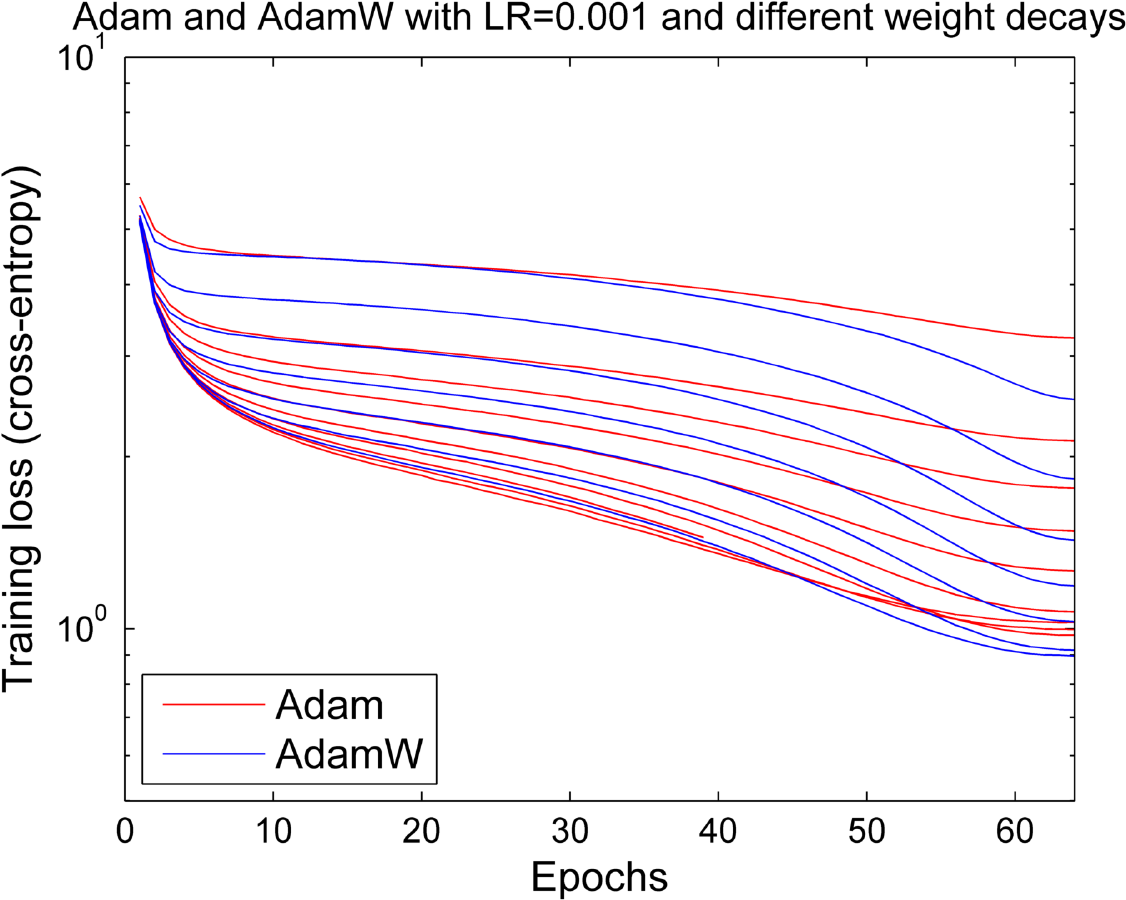} $\;\;$
  \includegraphics[width=0.47\textwidth]{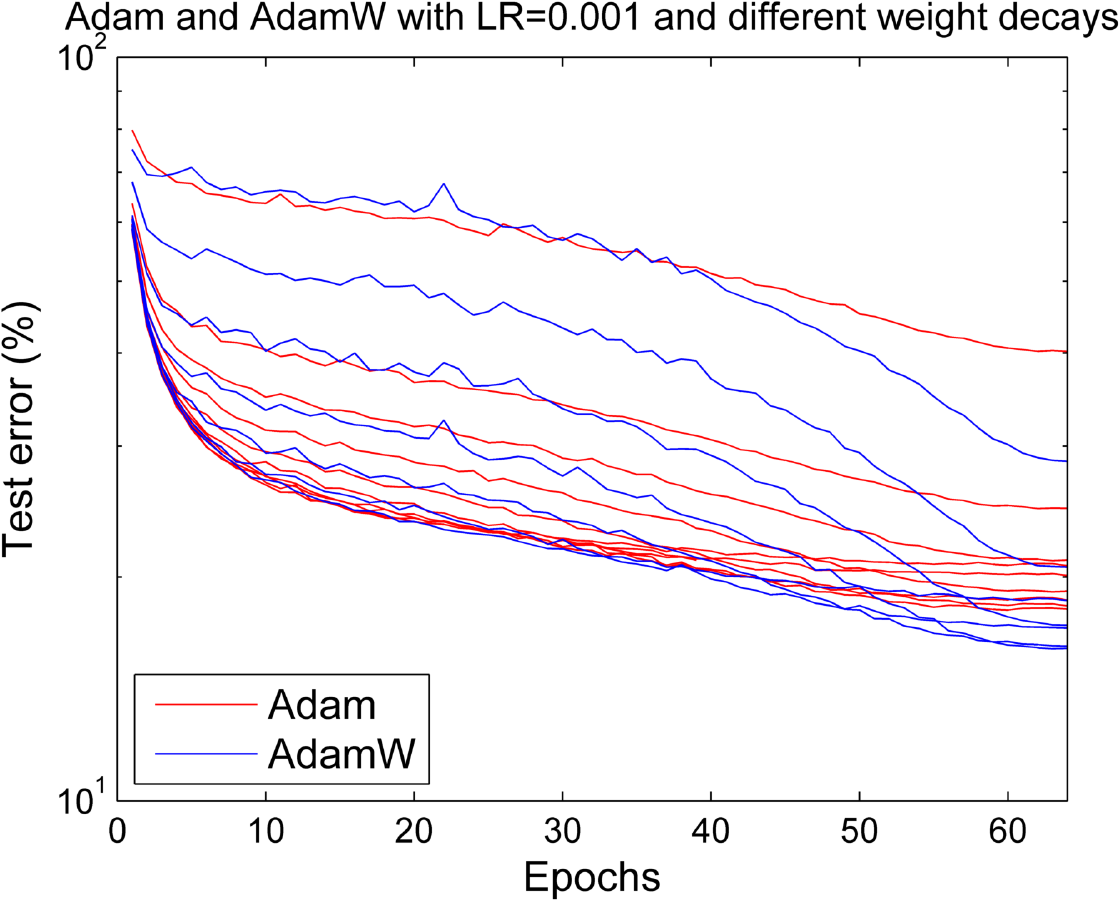}\\
	$\;\;$\\
	\includegraphics[width=0.47\textwidth]{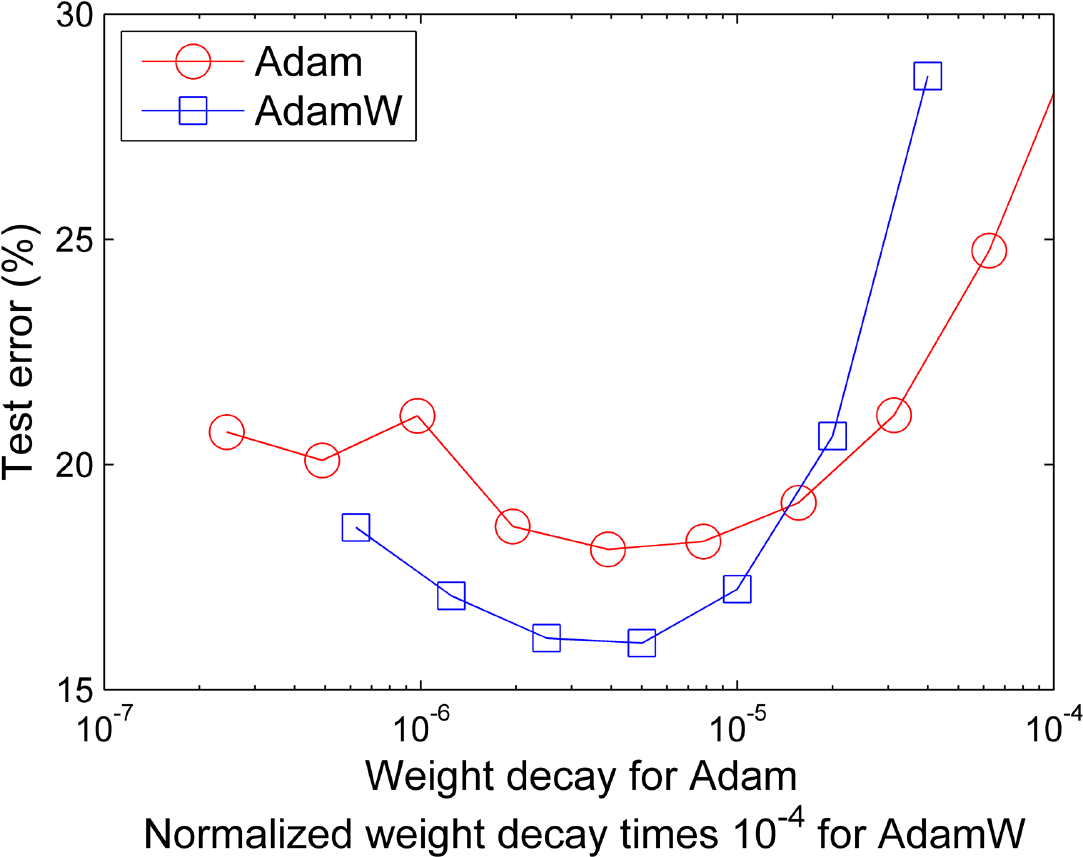} $\;\;$
  \includegraphics[width=0.47\textwidth]{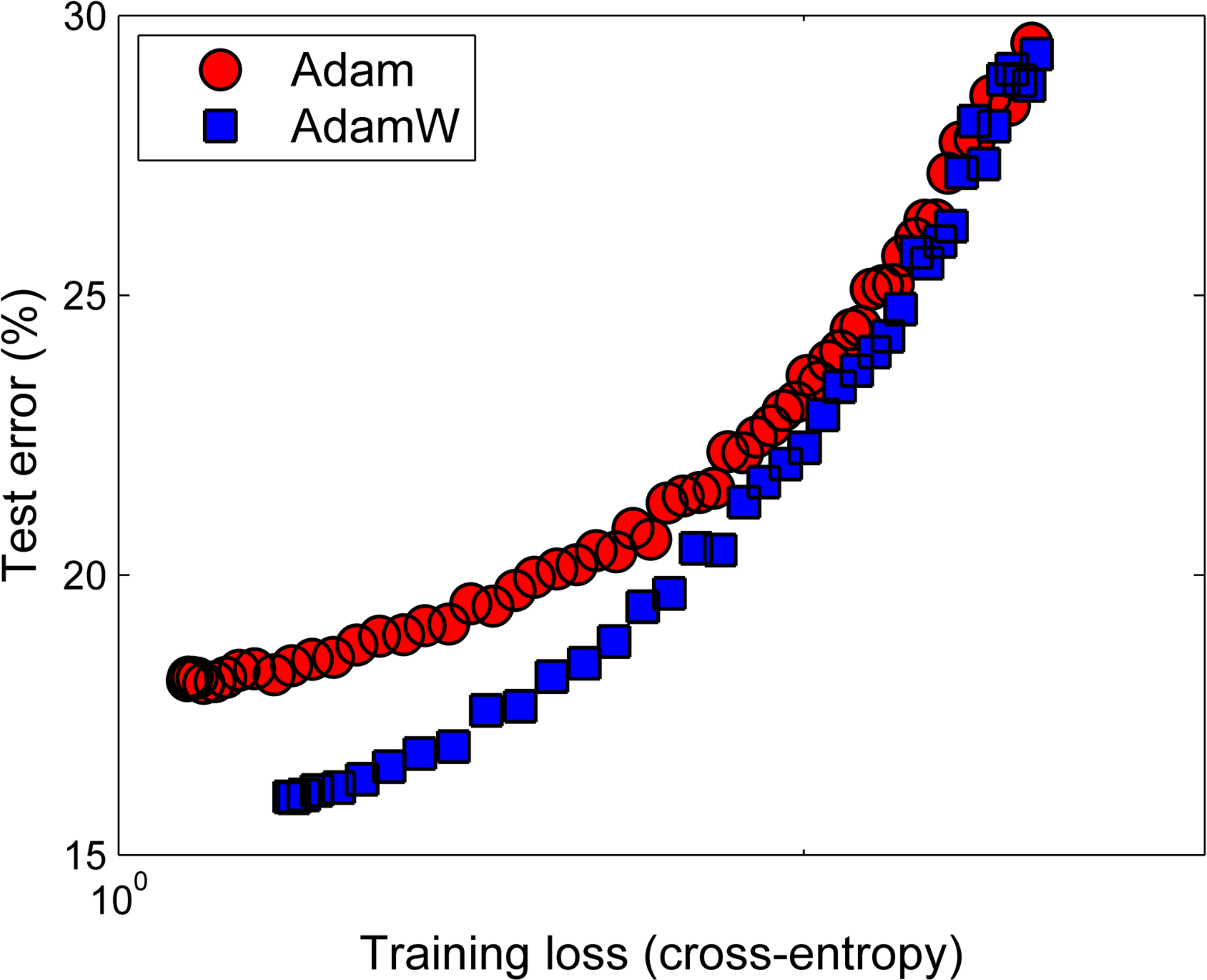}
\caption{\label{fig64ImageNet} Learning curves (top row) and generalization results (Top-5 errors in bottom row) obtained by a 26 2x96d ResNet trained with Adam and AdamW on ImageNet32x32.}
\end{figure*}

SuppFigure \ref{fig64ImageNet} is the equivalent of Figure 3 in the main paper, but for ImageNet32x32 instead of for CIFAR-10. The  qualitative results are identical: weight decay leads to better training loss (cross-entropy) than L$_2$ regularization, and to an even greater improvement of test error.

\begin{figure*}[t]%
	\includegraphics[width=0.9\textwidth]{restarts} \\
  \includegraphics[width=0.9\textwidth]{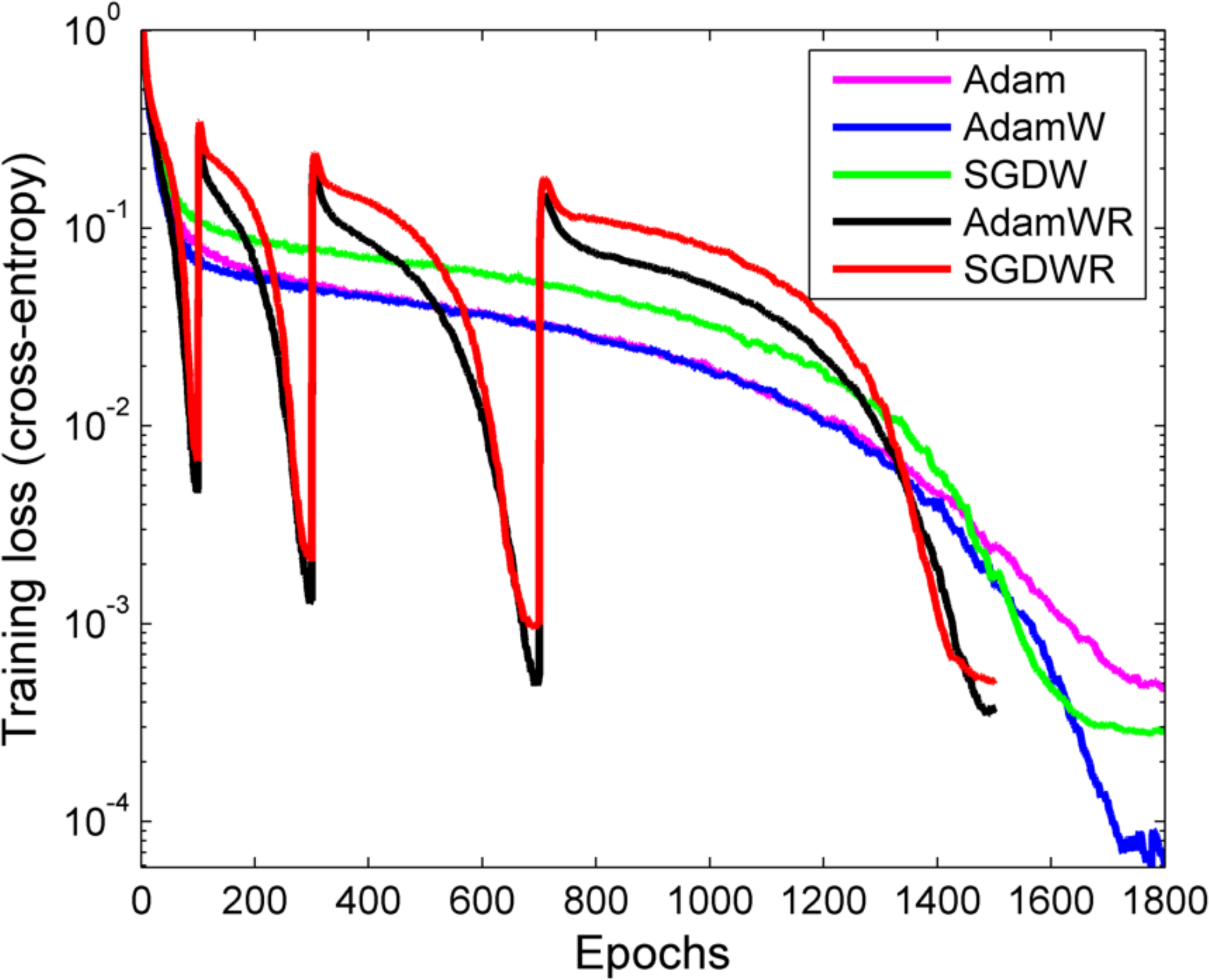}
\caption{\label{figtrainingCIFAR} Test error curves (top row) and training loss curves (bottom row) for CIFAR-10.}
\end{figure*}

\begin{figure*}[t]%
	\includegraphics[width=0.9\textwidth]{restartImageNet_zoom} \\
  \includegraphics[width=0.9\textwidth]{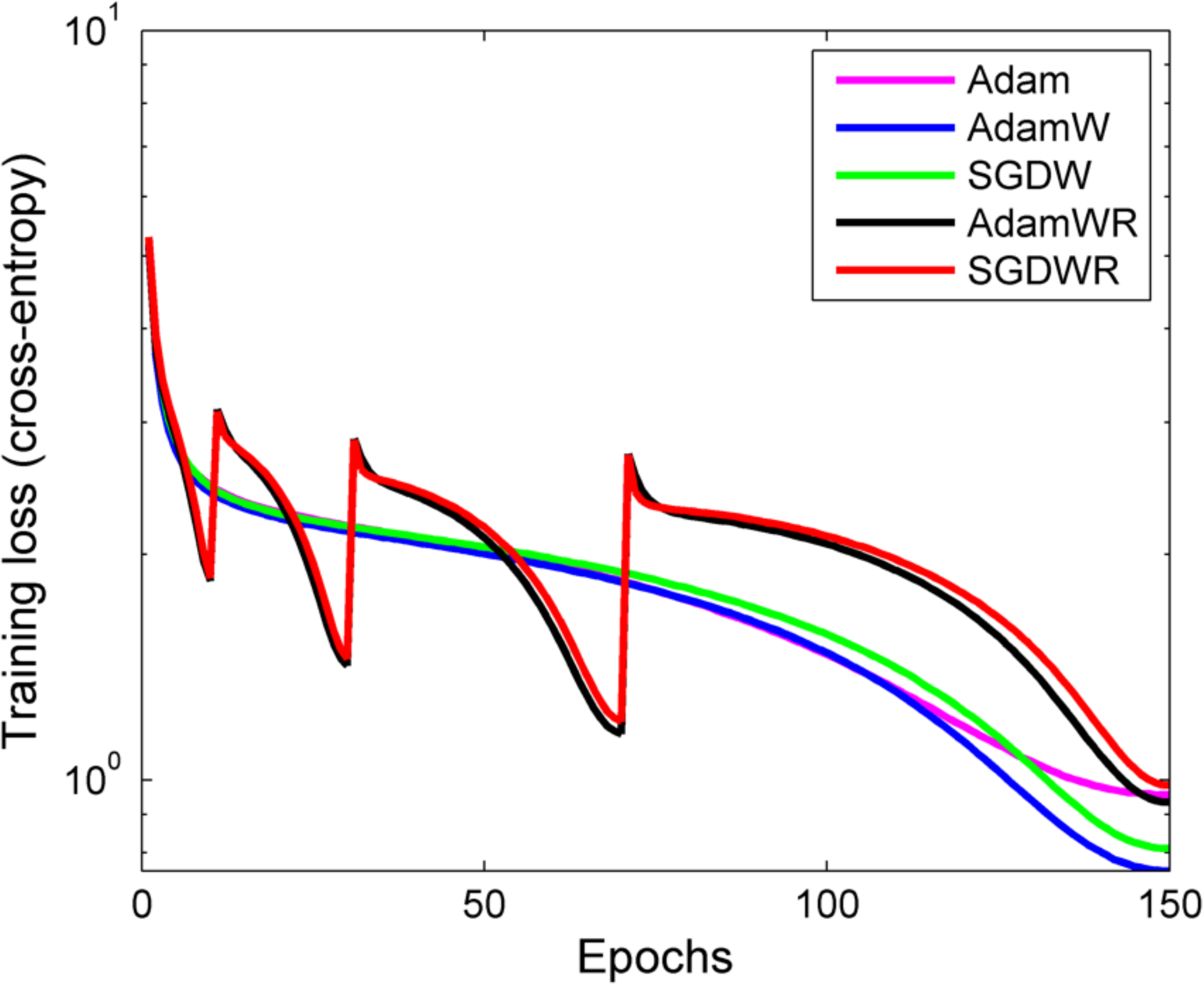}
\caption{\label{figtrainingImagenet32} Test error curves (top row) and training loss curves (bottom row) for ImageNet32x32.}
\end{figure*}

SuppFigure \ref{figtrainingCIFAR} and SuppFigure \ref{figtrainingImagenet32} are the equivalents of Figure 4 in the main paper but supplemented with training loss curves in its bottom row. The results show that Adam and its variants with decoupled weight decay converge faster (in terms of training loss) on CIFAR-10 than the corresponding SGD variants (the difference for ImageNet32x32 is small). As is discussed in the main paper, when the same values of training loss are considered, AdamW demonstrates better values of test error than Adam. Interestingly, SuppFigure \ref{figtrainingCIFAR} and SuppFigure \ref{figtrainingImagenet32} show that the restart variants AdamWR and SGDWR also demonstrate better generalization than AdamW and SGDW, respectively.

%\begin{figure*}[t]%
%	\includegraphics[width=0.47\textwidth]{restarts} $\;\;$
%  \includegraphics[width=0.47\textwidth]{restartImageNet_zoom}\\
%	$\;\;$\\
%	\includegraphics[width=0.47\textwidth]{cifar10_training} $\;\;$
%  \includegraphics[width=0.47\textwidth]{imagenet_training}
%\caption{\label{figtraining} Test error curves (top row) and training loss curves (bottom row) for CIFAR-10 (left column) and ImageNet32x32 (right column) datasets.}
%\end{figure*}

%\begin{figure*}[t]%
%	\includegraphics[width=0.99\textwidth]{BothAdam2.pdf}
%\caption{\label{figExternalResults} Learning curves (first and second columns) and generalization results (third and fourth columns) obtained by Adam and AdamW when trained on a Deep4Net architecture to classify and characterize the error-related brain response as measured in 24 intracranial 
%electroencephalography recordings (one per patient). The results were obtained and communicated to us by \cite{volker2018intracranial}. The results were obtained on their best performing problem-specific network architecture Deep4Net and with the same hyperparameter settings 
%}
%\end{figure*}

\end{document}